\documentclass[journal,twocolumn,10pt,compsoc]{IEEEtran}

\usepackage{times}
\usepackage{epsfig}
\usepackage{graphicx}
\usepackage{amsmath}
\usepackage{amssymb}

\usepackage{caption}
\usepackage[dvipsnames]{xcolor}
\usepackage{mathtools}
\DeclareMathOperator*{\argmax}{arg\,max}

\usepackage{courier}
\usepackage{makecell}
\usepackage{color, colortbl}
\definecolor{Gray}{gray}{0.5}
\usepackage{wrapfig}
\usepackage{float}
\usepackage{bbm}
\usepackage{pifont}
\usepackage{courier}
\usepackage{multirow}
\usepackage{amsfonts}
\usepackage{breakcites}
\usepackage{booktabs}
\usepackage{savesym}
\savesymbol{checkmark}
\usepackage{dingbat}

\newcommand{\cmark}{\ding{51}}
\newcommand{\xmark}{\ding{55}}

\newcommand{\norm}[1]{\left\lVert#1\right\rVert}

\definecolor{mygray}{gray}{0.9}

\newcommand{\eg}{\emph{e.g.}}
\newcommand{\ie}{\emph{i.e.}}
\newcommand{\etal}{\emph{et al.}} 

\ifCLASSOPTIONcompsoc
  \usepackage[nocompress]{cite}
\else
  \usepackage{cite}
\fi

\begin{document}

\title{Convolutional Hough Matching Networks\\ for Robust and Efficient Visual Correspondence}

\author{Juhong Min,
        Seungwook Kim,
        and~Minsu Cho
}

\markboth{}%
{Shell \MakeLowercase{\textit{et al.}}: Bare Demo of IEEEtran.cls for Computer Society Journals}

\IEEEtitleabstractindextext{%
\begin{abstract}
Despite advances in feature representation, leveraging geometric relations is crucial for establishing reliable visual correspondences under large variations of images. In this work we introduce a Hough transform perspective on convolutional matching and propose an effective geometric matching algorithm, dubbed Convolutional Hough Matching (CHM).
The method distributes similarities of candidate matches over a geometric transformation space and evaluates them in a convolutional manner. 
We cast it into a trainable neural layer with a semi-isotropic high-dimensional kernel, which learns non-rigid matching with a small number of interpretable parameters. 
To further improve the efficiency of high-dimensional voting, we also propose to use an efficient kernel decomposition with center-pivot neighbors, which significantly sparsifies the proposed semi-isotropic kernels without performance degradation.
To validate the proposed techniques, we develop the neural network with CHM layers that perform convolutional matching in the space of translation and scaling. 
Our method sets a new state of the art on standard benchmarks for semantic visual correspondence, proving its strong robustness to challenging intra-class variations.
\end{abstract}

\begin{IEEEkeywords}
Semantic visual correspondence, Hough matching, convolutional matching, center-pivot convolution
\end{IEEEkeywords}}

\maketitle
\IEEEdisplaynontitleabstractindextext
\IEEEpeerreviewmaketitle



\IEEEraisesectionheading{\section{Introduction}\label{sec:introduction}}

\IEEEPARstart{V}{isual} correspondence lies at the heart of image understanding, being a core component in numerous tasks such as object recognition, image retrieval, 
object tracking, and reconstruction~\cite{forsyth:hal-01063327}. 
With recent advances in deep neural networks \cite{he2016deep,hu2017senet,huang2015dense,krizhevsky2012imagenet,simonyan2015vgg}, there has been substantial progress in learning robust feature representation for establishing correspondences. 
Despite the effectiveness of deep convolutional features, however, spatial matching with a geometric constraint is still essential to handle image pairs with large variations, \eg, viewpoint and illumination changes.
In particular, the presence of intra-class variations, \ie, 
different instances of the same categories, remains a critical challenge
~\cite{
han2017scnet, huang2019dynamic, kim2017dctm, lee2019sfnet, liu2020semantic, min2019hyperpixel, min2020dhpf, rocco17geocnn, rocco18weak, rocco2018neighbourhood}.
The process of geometric matching is the de facto solution of choice, which most recent methods adopt in their models.  

Geometric matching commonly relies on exploiting a geometric consensus of candidate matches to verify relative transformations. In computer vision, RANSAC~\cite{fischler1981ransac} and Hough transform~\cite{hough1962transform} have long been used as geometric verification for wide-baseline correspondence problems with rigid motion models, while graph matching~\cite{cho2013learning, cho2010reweighted, fey2020deep, rolinek2020deep} has played a main role in matching deformable objects with  non-rigid motion. 
Recent work~\cite{cho2015unsupervised,han2017scnet,min2019hyperpixel,min2020dhpf} has advanced the idea of Hough transform to perform non-rigid image matching, showing that the Hough voting process incorporated in neural networks is effective for challenging correspondence problems with intra-class variations.
However, their matching modules are neither fully differentiable nor learnable, and weak to background clutter due to the position-invariant global Hough space.

\begin{figure}
    \begin{center}
        \includegraphics[width=1.0\linewidth]{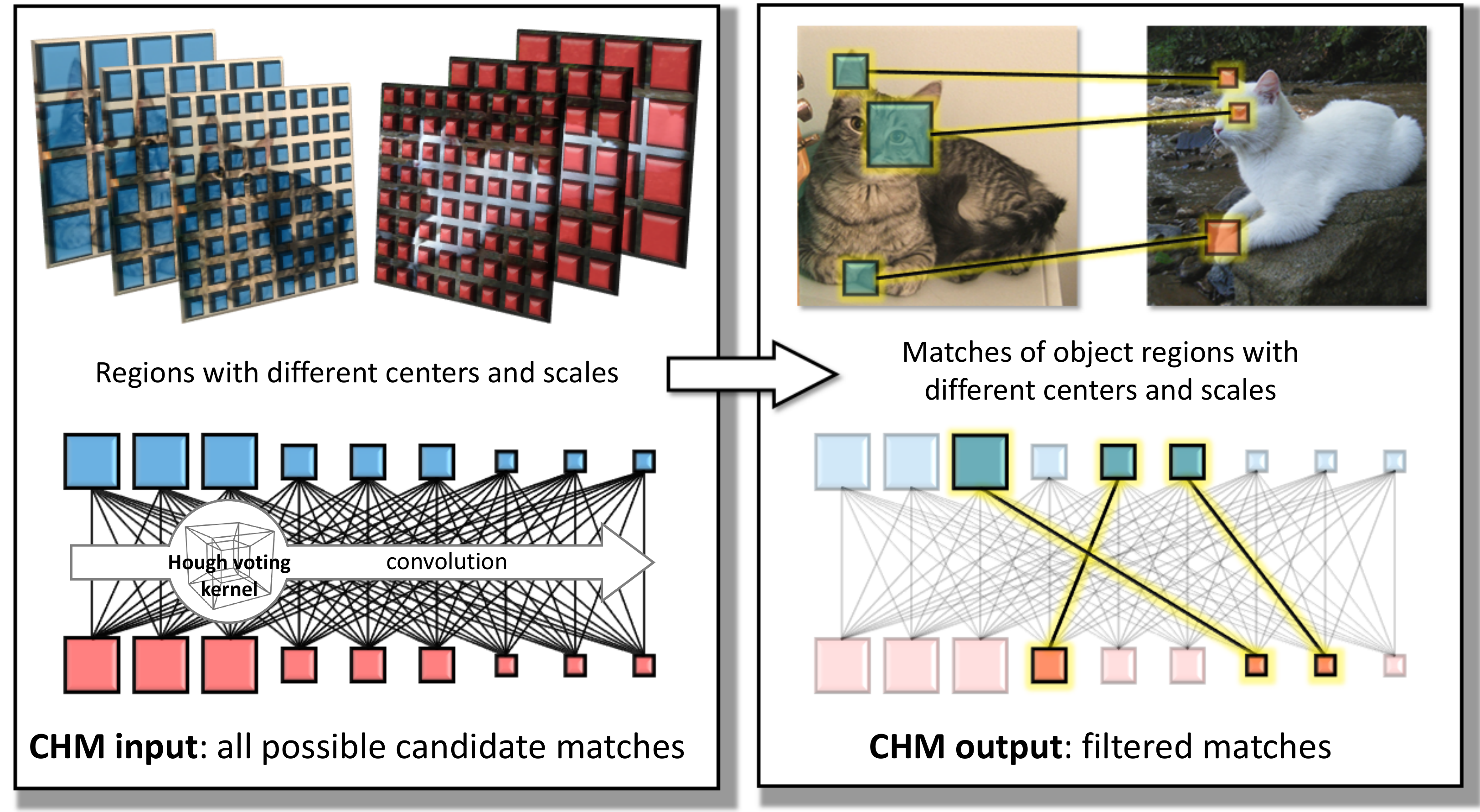}
    \end{center}
    \vspace{-3.0mm} 
      \caption{Convolutional Hough matching (CHM) establishes reliable correspondences across images by performing position-aware Hough voting in a high-dimensional geometric transformation space, \eg, translation and scaling.}
    \vspace{-7.0mm} 
\label{fig:teaser}
\end{figure}

In this work we introduce {\em Convolutional Hough Matching} (CHM) that distributes similarities of candidate matches over a geometric transformation space and evaluates them in a convolutional manner. As illustrated in Fig.~\ref{fig:teaser}, the convolutional nature makes the output equivariant to translation in the transformation space and also attentive to each position with its surrounding contexts, thus bringing robustness to background clutter. 
We design CHM as a learnable layer with an semi-isotropic high-dimensional kernel that acts on top of a correlation tensor. 
The CHM layer is compatible with any neural networks that use correlation computation, allowing flexible non-rigid matching and even multiple matching surfaces or objects.
It naturally generalizes existing 4D convolutions~\cite{huang2019dynamic, li2020correspondence, truong2020glunet, rocco2018neighbourhood} and provides a new perspective of Hough transform on convolutional matching. 
We also adapt the idea of center-pivot convolution~\cite{min2021hypercorrelation} to the semi-isotropic CHM kernels for efficient high-dimensional Hough voting with a linear complexity.
To demonstrate the proposed techniques, we propose the neural network with CHM layers that perform convolutional matching in the high-dimensional space of translation and scaling. 
Our method clearly outperforms state-of-the-art methods on standard benchmarks for semantic correspondence, proving its strong robustness to challenging intra-class variations.

This work is extended from the preliminary version of~\cite{min2021chm} by (1) introducing an effective kernel decomposition with center-pivot neighbors, (2) improving the performance with multi-level features and random data augmentation, and (3) providing additional experimental results and analyses.


\section{Related Work}

\smallbreak
\noindent \textbf{Hough transformation.}
The Hough transform~\cite{hough1962transform} is a classic method developed to identify primitive shapes in an image via geometric voting in a parameter space.
Ballard~\cite{ballard1981generalhough} generalizes the idea to identify positions of arbitrary shapes with R-table.
Early approaches~\cite{chen2013robust, cho2012progressive} in computer vision widely adopt Hough transform for its effectiveness in extracting features of a particular shape in an image.
As a representative example, Leibe \etal\cite{leibe2003interleaved} introduce a Hough-based object segmentation and detection method by incorporating information about supporting patterns of parts for the target category. 
The idea of Hough voting has widely been adopted in diverse tasks including retrieval~\cite{huan2017vehicle}, object discovery~\cite{Gall2009, MILLETARI201792, novotny2018semi, qi2019deep}, shape recovery~\cite{sun2010depth}, 3D vision~\cite{Knopp2011SceneCC, jan2010orientation}, and pose estimation~\cite{kehl2016eccv} to name a few. In geometric matching, Cho \etal\cite{cho2015unsupervised} first extends it to the Probabilistic Hough Matching (PHM) algorithm for unsupervised object discovery.
Recent methods~\cite{ham2016proposal,ham2018proposal,han2017scnet,kawk2015unsupervised,liu2020semantic,min2019hyperpixel,min2020dhpf,sultani2016what} have demonstrated the efficacy of the Hough matching with good empirical performance. 
They, however, are all limited in the sense that the geometric voting is carried out to discover a {\em global} offset consensus rather than a {\em local and individual} consensus for a match, which makes it less accurate and weak to clutter.

\smallbreak
\noindent \textbf{Semantic visual correspondence.}
Traditional approaches to semantic correspondence ~\cite{bristow2015dense, cho2015unsupervised, ham2016proposal, ham2018proposal, kim2013deformable, liu2011sift, taniai2016joint, yang2017object} typically use hand-crafted descriptors~\cite{bay2006surf, dalal2005histograms, lowe2004sift}.
Although the classic methods work satisfactorily for some applications, they still suffer apparent disadvantages \eg, lack of semantic patterns.
Recent approaches~\cite{han2017scnet, jeon2018parn, jeon2020guided, liu2020semantic, min2019hyperpixel, min2020dhpf, rocco17geocnn, rocco18weak, rocco2018neighbourhood, paul2018attentive, truong2020glunet, wang2019learning} build upon features from convolutional neural network (CNN) pretrained on classification task~\cite{deng2009imagenet}.
Han~\etal\cite{han2017scnet} introduce a CNN-based matching model that learns to compute a correlation tensor.
Min~\etal\cite{min2019hyperpixel, min2020dhpf, min2021hypercorrelation} show that exploiting multi-layer CNN features is beneficial for establishing fine-grained semantic visual correspondences, offering useful insights in extending our method.
Rocco~\etal\cite{rocco17geocnn} propose to learn a CNN regressor that computes a series of 2D convolutions on a dense correlation matrix to predict global geometric transformation parameters, either affine or TPS~\cite{donato2002approximate}.
Seo~\etal\cite{paul2018attentive} improve the framework with offset-aware correlation kernels with attention modules.
Jeon~\etal\cite{jeon2018parn} stack multiple affine transformation networks and compute correspondences in coarse-to-fine manner.
Wang~\etal\cite{wang2019learning} adopt the CNN architecture to estimate translation and rotation parameters to learn correspondences from raw video.
These methods demonstrate that a series of 2D convolutions acting on correlation tensors is effective in capturing geometric information by exploiting local patterns of similarity. 

\smallbreak
\noindent \textbf{4D convolution for visual correspondence.}
Rocco~\etal\cite{rocco2018neighbourhood} introduce the neighbourhood consensus network that uses 4D convolution for visual correspondence. They view 4D convolution as an extension of 2D convolution, which learns multiple similarity patterns of local correspondences, and thus use multiple 4D kernels, requiring a large number of parameters to learn. 
Following the work, recent methods~\cite{huang2019dynamic, li2020correspondence, rocco2018neighbourhood, truong2020glunet} also adopt 4D convolution in a similar manner. They commonly consume a high computational cost with a large number of parameters in the kernels and only consider translation in space. 
In contrast, we extend the idea of Hough matching~\cite{cho2015unsupervised} for high-dimensional convolution and propose an interpretable and light-weight (semi-isotropic) high-dimensional kernel for visual correspondence. In doing so, it naturally generalizes the existing 4D convolution to higher dimensions and achieves superior performance using only a single kernel per layer with a small number of parameters. The results reveal that the role of high-dimensional convolution on a correlation tensor for matching is to learn a reliable voting strategy rather than to capture diverse patterns in the correlation tensor.

\smallbreak
\noindent \textbf{Efficient convolutional networks.}
Recently, there has been increasing interests in designing deep convolutional networks with less computation and memory requirements.
Depth-wise separable convolutions~\cite{mobilev2_2018,chollet2017depthconv,howard2017mobilenets,zhang2018shuffle} separate a multi-channel 2D convolution into a depth-wise convolution and a point-wise convolution.
Work of~\cite{jaderberg2014speeding, Lebedev2015SpeedingupCN} propose to speed up training via tensor factorization.
U-Net encoder-decoder structures are also widely used~\cite{badrina2017segnet,dosovitskiy2015flownet,Ronneberger2015unet}, where they downsample the input feature maps with strided convolutions and upsample them back for lower memory and computation.
Yang \etal~\cite{yang2019optical} separates a 4D convolution into two separate 2D kernels, and downsamples the 4D cost-volume to maintain small memory footprints.
In this study, we propose a learnable convolutional layer using position-sensitive isotropic kernel, dubbed CHM. 
Furthermore, inspired by the recent work of~\cite{min2021hypercorrelation} that mitigate the quadratic complexity of high-dimensional convolutions via effective kernel decomposition, we introduce center-pivot neighbors to our semi-isotropic high-dimensional CHM kernels, bringing significant improvements in terms of both memory and time.

Our contributions can be summarized as follows:\vspace{-5px}
\begin{itemize}
\setlength\itemsep{0em}
    \item We introduce a Hough transform perspective on convolutional matching and propose an effective geometric matching algorithm, CHM, which performs high-dimensional Hough voting in a convolutional manner.
    \item We develop CHM into a trainable neural layer with a semi-isotropic high-dimensional kernel, which learns non-rigid matching with a small number of interpretable parameters.
    \item We improve the efficiency of high-dimensional Hough voting of CHM by applying the idea of center-pivot convolution to our semi-isotropic CHM kernels.
    \item We propose the convolutional Hough matching network (CHMNet) that performs geometric matching in a translation and scaling space using 6D convolution. 
    \item The proposed method sets a new state of the art on standard benchmarks for semantic visual correspondence, proving its robustness to challenging intra-class variations across images to match.
\end{itemize}

\section{Convolutional Hough Matching}\label{sec:CHM}

In this section, we revisit the Hough matching method for visual correspondence and then propose its convolutional version as a high-dimension convolutional layer, which is readily trainable in neural networks. 

\subsection{Hough matching \& its convolutional extension}

The Hough transform is a powerful detection method for a geometric object, which exploits the duality between parts and parameters of the object~\cite{ballard1981generalhough, hough1962transform}. It performs voting in a parameter space of the target object, called the Hough space, where votes from the object parts are accumulated to form local maxima in the space. The objects are then detected simply by identifying the positions of local maxima. 
The Hough matching method~\cite{cho2015unsupervised}, inspired by the Hough transform, detects reliable correspondences by geometric voting from candidate matches. Given two images, it constructs the Hough space of parameters of geometric transformation between the two images and then accumulates votes of candidate matches for plausible transformation. 

Let us assume a local region $\mathbf{x}$ on an image, that is represented by its geometric attributes, \ie, pose and shape. In principle $\mathbf{x}$ can be a form of any parameterization, but in this work we simply describe the region $\mathbf{x}$ by its center and scale.   
Now let us consider two images, $I$ and $I'$, and two sets of local regions, $\mathcal{X}$ and $\mathcal{X}'$, obtained from the two images, respectively. For any two regions $(\mathbf{x}, \mathbf{x}') \in \mathcal{X} \times \mathcal{X}'$, a correlation function $c$ computes a non-negative similarity $c(\mathbf{x}, \mathbf{x}')$ using appearance features of the regions. The main idea of Hough matching is to create the Hough space $\mathcal{H}$, that is the space of all possible offsets $\mathbf{h}$ between two regions, \ie, translation and scaling, and accumulate votes from candidate matches onto the Hough space as
\begin{align}
    v(\mathbf{h}) =   \sum_{ (\mathbf{x}, \mathbf{x}') \in \mathcal{X} \times \mathcal{X}'} c(\mathbf{x}, \mathbf{x}') k_\text{iso}( \| (\mathbf{x}' - \mathbf{x}) - \mathbf{h} \|_\mathrm{g} ),  
\end{align}
where $ \|\cdot\|_\mathrm{g}$ represents a group-wise distance function that computes the distances separately for two groups, center and scale, \ie, $ \| \mathbf{x} \|_\mathrm{g} = [ \| \mathbf{x}_\mathrm{xy} \|; \|  \mathbf{x}_\mathrm{s} \| ]$ (subscripts $\mathrm{xy}$ for center and $\mathrm{s}$ for scale) and $k_\text{iso}$ is a kernel function that computes similarity between the observed offset, $\mathbf{x}' - \mathbf{x}$, and the given offset $\mathbf{h}$ in the Hough space.\footnote{For the kernel function, previous work uses a form of discretized Gaussian~\cite{cho2015unsupervised} or Dirac delta~\cite{han2017scnet} without learning the kernel parameters.} The kernel $k_\text{iso}$ is designed to assign a voting weight for each candidate match according to how close the offset induced by the match $(\mathbf{x}, \mathbf{x}')$ is to $\mathbf{h}$; we use the group-wise distance to differentiate the effects of center and scale in the kernel. 
The resultant voting map $v(\mathbf{h})$ over the Hough space $\mathcal{H}$ can be used to find reliable matches $(\mathbf{x}, \mathbf{x}')$ by suppressing spurious ones corresponding to relatively low voting scores $v(\mathbf{x}'-\mathbf{x})$, \eg, updating the match score via $c(\mathbf{x}, \mathbf{x}') v(\mathbf{x}'-\mathbf{x})$~\cite{han2017scnet}.
Despite its good empirical performance~\cite{cho2015unsupervised,ham2016proposal,ham2018proposal,han2017scnet,kawk2015unsupervised,liu2020semantic,min2019hyperpixel,min2020dhpf,sultani2016what}, the global voting map $v(\mathbf{h})$, which is shared for all candidate matches, is limited in the sense that it cannot capture the reliability of a specific candidate match. This global position-invariant Hough space makes the output less accurate and weak to background clutter, \eg, increasing the score of distant outliers that has a similar offset to that of dominant inliers.   

\begin{figure}[t]
    \begin{center}
        \includegraphics[width=1.0\linewidth]{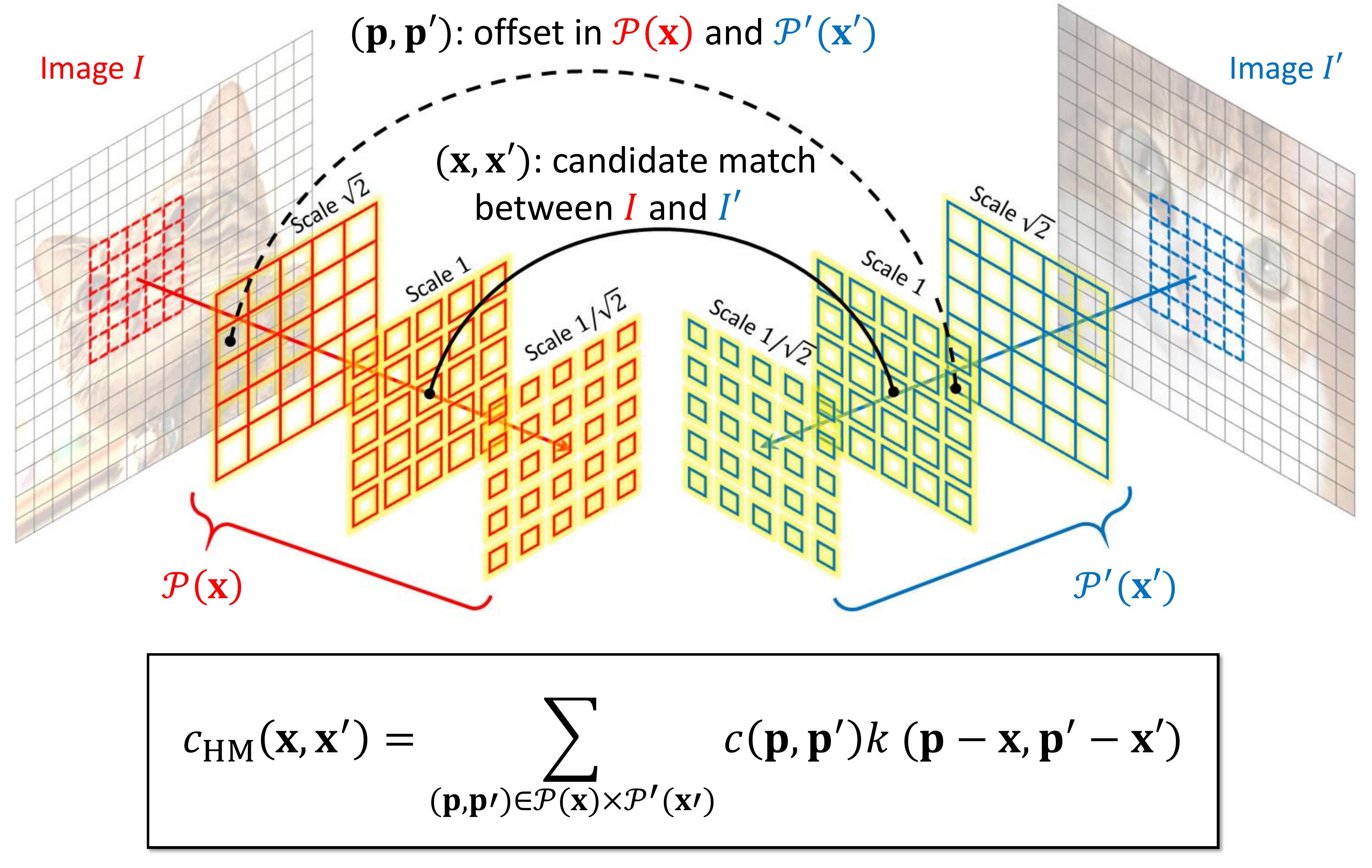}
    \end{center}
    \vspace{-3.0mm}
    \caption{Convolutional Hough matching that carries out geometric voting in 6D space, \eg, translation and scale.}\label{fig:CHM}
    \vspace{-5.0mm}
\end{figure}

As illustrated in Fig.~\ref{fig:CHM}, in order to address the issue, we create a local and individual voting space for each candidate match $(\mathbf{x}, \mathbf{x}')$ by introducing local windows around the regions, $\mathbf{x}$ and $\mathbf{x}'$:
\begin{align}
    v(\mathbf{x}, \mathbf{x}', \mathbf{h}) = \!\!\!\! \sum_{ (\mathbf{p}, \mathbf{p}') \in \mathcal{P}(\mathbf{x}) \times \mathcal{P}'(\mathbf{x'})} \!\!\! c(\mathbf{p}, \mathbf{p}') k_\text{iso}( \| (\mathbf{p}' - \mathbf{p}) - \mathbf{h} \|_\mathrm{g} ), 
\end{align} 
where $\mathcal{P}(\mathbf{x})$ denotes the set of neighbor regions within the local window centered on $\mathbf{x}$. 
Since this local voting space is now dedicated to $(\mathbf{x}, \mathbf{x}')$, 
we can simply assign a match score $v$ for the candidate match by taking the vote value at the bin with offset zero:  
\begin{align}
    \label{eqn:houghmatching_1} 
   v(\mathbf{x}, \mathbf{x}') &= \sum_{ (\mathbf{p}, \mathbf{p}') \in \mathcal{P}(\mathbf{x}) \times \mathcal{P}'(\mathbf{x'})} c(\mathbf{p}, \mathbf{p}') k_\text{iso}( \| \mathbf{p}' - \mathbf{p} \|_\mathrm{g} ). 
\end{align}
With a slight abuse of notation, let us use $k( \mathbf{z}, \mathbf{z}')$ to represent the kernel value corresponding to two positions, $\mathbf{z}$ and $\mathbf{z}'$, each representing a local region in the parameter space of regions, \ie, 3D space of center and scale in our case. The equation above then can be generalized to a form of 6D convolution with an arbitrary kernel $k$: 
\begin{align} 
   c_{\text{HM}}(\mathbf{x}, \mathbf{x}') &= \sum_{ (\mathbf{p}, \mathbf{p}') \in \mathcal{P}(\mathbf{x}) \times \mathcal{P}'(\mathbf{x}')} c(\mathbf{p}, \mathbf{p}') k( \mathbf{p} - \mathbf{x}, \mathbf{p}' - \mathbf{x}') \nonumber  \\ &= (c \ast k) (\mathbf{x}, \mathbf{x}'), 
\end{align}
which becomes equivalent to Eq.~\ref{eqn:houghmatching_1} when the group-wise isotropic kernel $ k_\text{iso}$ is used. 

Note that this convolutional extension of Hough matching has a generic form; it reduces to a similar form of 4D convolutions in~\cite{huang2019dynamic, li2020correspondence, rocco2018neighbourhood, truong2020glunet} when the Hough space is restricted to center translation, and generalizes to higher dimensions beyond 6D when additional transformation dimensions is introduced such as rotation, shear, and others.

\subsection{Convolutional Hough matching layer}
\label{sec:CHM_layer}
We design the convolutional Hough matching (CHM) as a learnable convolution layer: 
\begin{align}
   c_{\text{HM}}(\mathbf{x}, \mathbf{x}'; k, b) = b +  (c \ast k) (\mathbf{x}, \mathbf{x}'), 
\end{align}
where $b$ is a bias term for the layer and $k$ represents a kernel with a specific type of weight sharing. 
The group-wise isotropic kernel $k_{\text{iso}}$, which is directly derived from Hough matching, can be implemented by weight sharing among parameters with the same offset $|\mathbf{z}-\mathbf{z}'|$ in $k( \mathbf{z}, \mathbf{z}')$. 
While it is a reasonable choice, the fully isotropic kernel assigns the same importance to the matches of the same offset regardless of their distances from the kernel position $(\mathbf{x}, \mathbf{x}')$. It may be an excessive constraint in the sense that the distance of an object from the center of focus is likely to be relevant to the importance. 

We thus relax the isotropy and propose the position-sensitive isotropic kernel $k_\text{psi}( \|\mathbf{p}' - \mathbf{p}\|_\mathrm{g}; \|\mathbf{p} - \mathbf{x}\|_\mathrm{g}, \|\mathbf{p}' - \mathbf{x}'\|_\mathrm{g} )$ that differentiates the distances from the kernel position, $\|\mathbf{p} - \mathbf{x}\|_\mathrm{g}$ and $\|\mathbf{p}' - \mathbf{x}'\|_\mathrm{g}$. The kernel $k_{\text{psi}}$ is implemented by sharing parameters whose triplets, $(\|\mathbf{p}' - \mathbf{p}\|_\mathrm{g}, \|\mathbf{p} - \mathbf{x}\|_\mathrm{g}, \|\mathbf{p}' - \mathbf{x}'\|_\mathrm{g})$, are the same. 

The CHM layer is compatible with any neural network layer that computes correlations between images, and can be stacked multiple times to improve the performance. As a result of substantial parameter sharing, the 6D kernels, $k_{\text{iso}}^{\mathrm{6D}}$ and $k_{\text{psi}}^{\mathrm{6D}}$ in $\mathbb{R}^{H_\mathrm{k} \times W_\mathrm{k} \times S_\mathrm{k} \times H_\mathrm{k} \times W_\mathrm{k} \times S_\mathrm{k}}$, contain only a small number of parameters, thus making CHM resistant to overfitting in training; \eg, the kernels with $H_\mathrm{k}=W_\mathrm{k}=5$ and $S_\mathrm{k}=3$ contains only 45 and 220 parameters, respectively, while the full kernel has 5,625. 
More importantly, the perspective of Hough matching on convolution provides the interpretability of the learned kernel: each element in the kernel is a voting weight of the corresponding neighbor match in the local offset space. 

\begin{figure}[t]
    \begin{center}
        \includegraphics[width=1.0\linewidth]{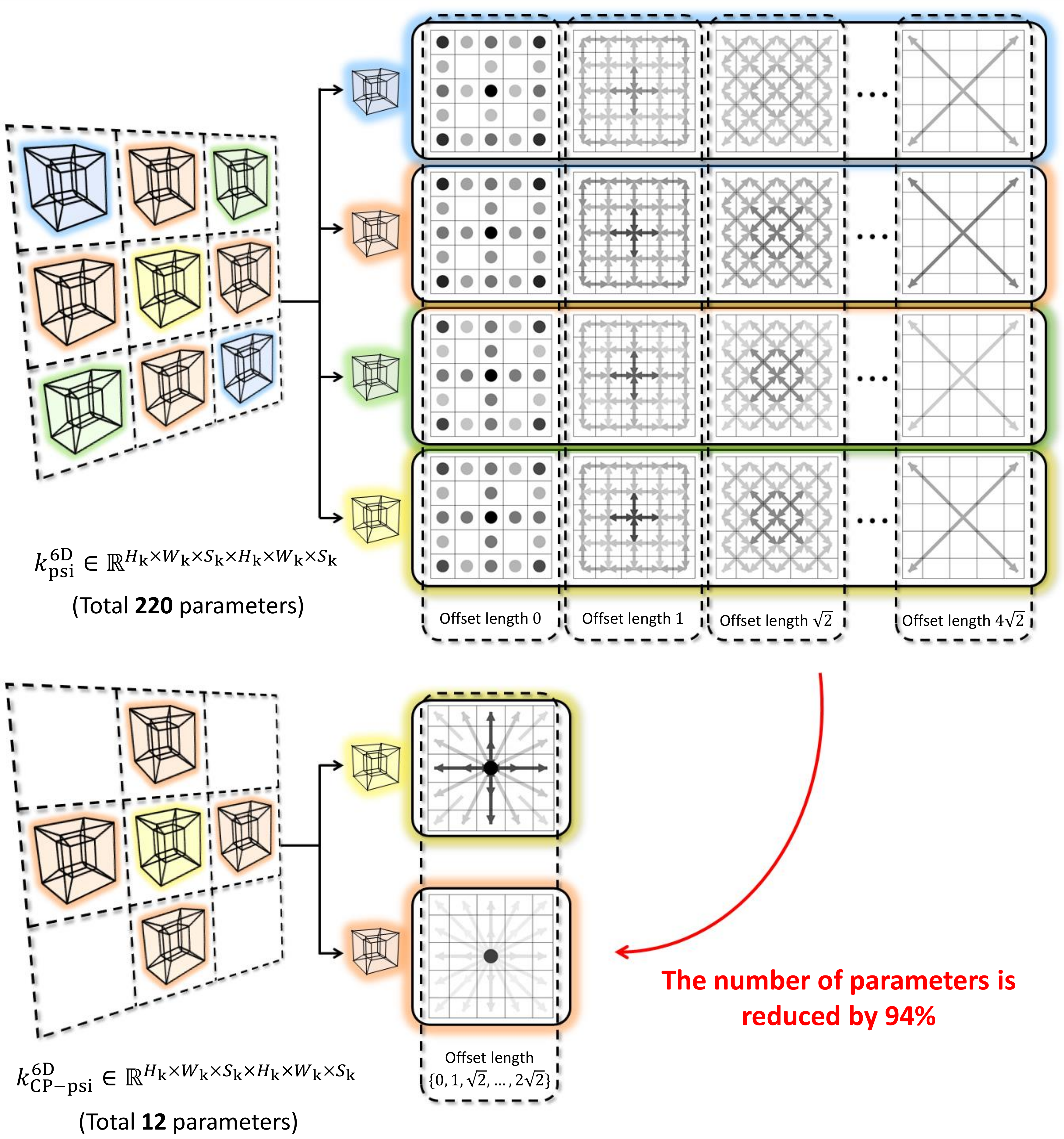}
    \end{center}
    \vspace{-3.0mm}
    \caption{Visualization of learned 6-dimensional CHM kernel $k_{\text{psi}}^{\mathrm{6D}}$ (top) and CP-CHM kernel $k_{\text{CP-psi}}^{\mathrm{6D}}$ (bottom) where $H_{\mathrm{k}} = W_{\mathrm{k}} = 5$ and $S_{\mathrm{k}} = 3$. See Fig.~\ref{fig:kernel_vis_demo} to see how we visualized them.}\label{fig:CHM_kernels}
    \vspace{-6.0mm}
\end{figure}

\begin{figure*}
    \begin{center}
        \includegraphics[width=1.0\linewidth]{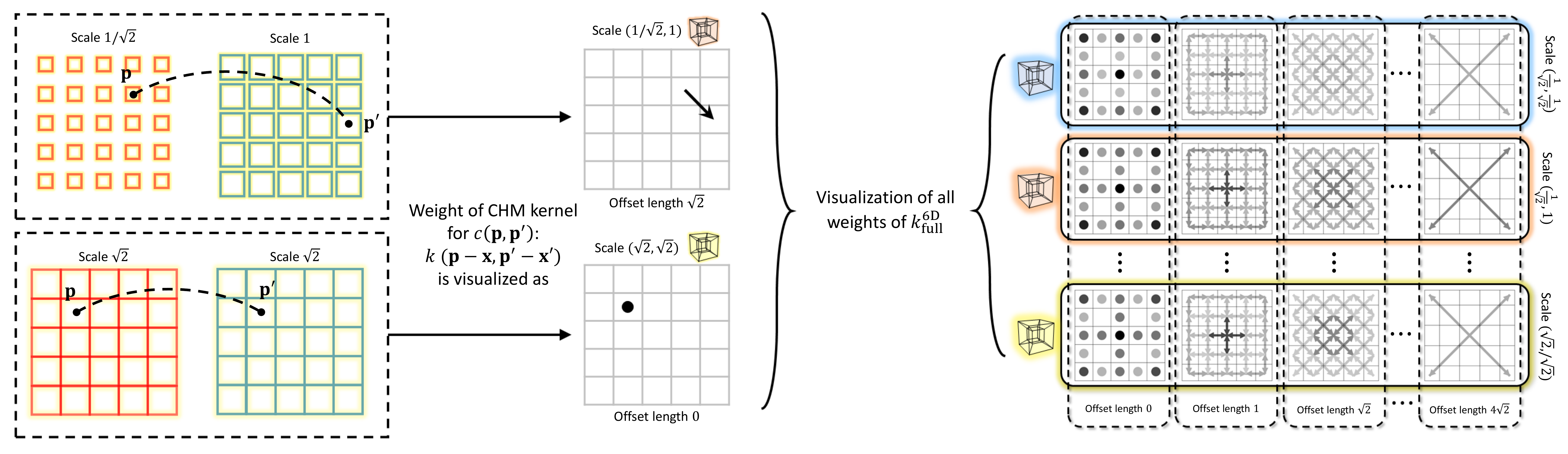}
    \end{center}
    \vspace{-5.0mm} 
       \caption{Description of visualizing learned weights of high-dimensional kernels: (Left) The arrows represent the offset vectors relative to the kernel position $(\mathbf{x}, \mathbf{x}')$, and the circles mean zero offset. (Right) For straightforward visualization, we decompose a high-dimensional kernel into multiple 4D kernels (tesseracts) and visualize learned weights of each 4D kernel as a set of maps consisting of offset vectors. Darker offsets mean larger weights while brighter ones mean smaller weights.}
    \vspace{-3.0mm} 
\label{fig:kernel_vis_demo}
\end{figure*}

\subsection{Center-pivot convolutional Hough matching layer}
The proposed Hough matching perspective on high-dimensional conv matching provides a number of benefits including interpretability of learned kernels, and generalizibility to higher dimensions, and scalability via parameter sharing.
However, the quadratic complexity with respect to the spatial size of the input tensor, \ie, the curse of dimensionality, still remains as an apparent bottleneck in terms of efficiency.
Inspired by the work of~\cite{min2021hypercorrelation} for few-shot segmentation, we introduce {\em center-pivot} neighbors to the $n$-dimensional CHM kernels.
The center-pivot neighbors in~\cite{min2021hypercorrelation} aims at reducing quadratic complexity of 4D convolutions down to linear complexity by disregarding a large number of insignificant neighbors in local 4D space, which eventually leads to an effective decomposition of a 4D kernel into a pair of 2D kernels.
In the rest of this subsection, we show such decomposition can be performed on any arbitrary $n$-dimensional CHM kernels.

Let $\mathcal{P}^{n\text{D}}(\mathbf{x}, \mathbf{x}') = \mathcal{P}(\mathbf{x}) \times \mathcal{P}'(\mathbf{x}')$ where each of $\mathcal{P}$ and $\mathcal{P}'$ returns a set of neighboring positions around the center in $\frac{n}{2}$-dimensional subspace, \eg, $\mathbf{x}, \mathbf{x}' \in \mathbb{R}^{\frac{n}{2}}$.
Instead of using every possible neighbors in the local $n$D window of interest, we collect only a small number of vital neighbors that pivot the given centers, $\mathbf{x}$ and $\mathbf{x}'$, while ignoring the others. 
Specifically, the set of center-pivot neighbors is defined as
\begin{align}
    \mathcal{P}^{n\text{D}}_{\text{CP}}(\mathbf{x}, \mathbf{x}') &= \mathcal{P}^{n\text{D}}_{c}(\mathbf{x}, \mathbf{x}') \cup \mathcal{P}^{n\text{D}}_{c'}(\mathbf{x}, \mathbf{x}') \\ \nonumber
    &= \{(\mathbf{p}, \mathbf{p}') \in \mathcal{P}(\mathbf{x}, \mathbf{x}'): \mathbf{p} = \mathbf{x}\} \\ \nonumber
    & \ \ \cup \{(\mathbf{p}, \mathbf{p}') \in \mathcal{P}(\mathbf{x}, \mathbf{x}'): \mathbf{p}' = \mathbf{x}'\}.
\end{align}
Let $k_{c}$ and $k_{c'}$ denote $n$D kernels with respective neighbors of $\mathcal{P}^{n\text{D}}_{c}(\mathbf{x}, \mathbf{x}')$ and $\mathcal{P}^{n\text{D}}_{c'}(\mathbf{x}, \mathbf{x}')$. Then, the center-pivot $n$D conv is formulated as a union of two separate convolutions:
\begin{align}
\label{eqn:cp-chm}
   c_{\text{CP}}(\mathbf{x}, \mathbf{x}') &= (c * k_{c})(\mathbf{x}, \mathbf{x}') + (c * k_{c'})(\mathbf{x}, \mathbf{x}').
\end{align}
Given $n$D kernel $k$, consider the first term in the right-hand side of Eqn.~\ref{eqn:cp-chm}:
\begin{align}
   (c * k_{c})(\mathbf{x}, \mathbf{x}') &= \sum_{(\mathbf{p}, \mathbf{p}') \in \mathcal{P}^{n\text{D}}_{c}(\mathbf{x}, \mathbf{x}')}c(\mathbf{p}, \mathbf{p}')k(\mathbf{p} - \mathbf{x}, \mathbf{p}' - \mathbf{x}') \\ \nonumber
   &= \sum_{\mathbf{p}' \in \mathcal{P}'(\mathbf{x}')}c(\mathbf{x}, \mathbf{p}')k(\mathbf{0}, \mathbf{p}' - \mathbf{x}') \\ \nonumber
   &= \sum_{\mathbf{p}' \in \mathcal{P}'(\mathbf{x}')}c(\mathbf{x}, \mathbf{p}')k^{\frac{n}{2}\text{D}}_{c}(\mathbf{p}' - \mathbf{x}'),
\end{align}
where $k^{\frac{n}{2}\text{D}}_{c} = k(\mathbf{0}, :)$ that convolves on $\frac{n}{2}$-dimensional subspace of the input tensor: $c(\mathbf{x}, :) * k(\mathbf{0}, :)$.
Similarly, we have
\begin{align}
   (c * k_{c'})(\mathbf{x}, \mathbf{x}') &= \sum_{\mathbf{p} \in \mathcal{P}(\mathbf{x})}c(\mathbf{p}, \mathbf{x}')k^{\frac{n}{2}\text{D}}_{c'}(\mathbf{p} - \mathbf{x}),
\end{align}
where $k^{\frac{n}{2}\text{D}}_{c'} = k(:, \mathbf{0})$.
The center-pivot high-dimensional convolution performs two separate convolutions on their corresponding subspaces, having a linear complexity with respect to the spatial size of the input:
\begin{align}
\label{eqn:CP_CHM}
   (c * k_{\text{CP}})(\mathbf{x}, \mathbf{x}') &= \sum_{\mathbf{p}' \in \mathcal{P}'(\mathbf{x}')}c(\mathbf{x}, \mathbf{p}')k^{\frac{n}{2}\text{D}}_{c}(\mathbf{p}' - \mathbf{x}')
    \\ \nonumber
    &+ \sum_{\mathbf{p} \in \mathcal{P}(\mathbf{x})}c(\mathbf{p}, \mathbf{x}')k^{\frac{n}{2}\text{D}}_{c'}(\mathbf{p} - \mathbf{x}),
\end{align}
where $k^{\frac{n}{2}\text{D}}_{c}, k^{\frac{n}{2}\text{D}}_{c'} \in \mathbb{R}^{H_\mathrm{k} \times W_\mathrm{k} \times S_\mathrm{k}}$ are two different 3D kernels in case of 6D convolution ($n=6$).

From the perspective of Hough matching, we further adapt different parameter sharing strategies discussed in Sec.~\ref{sec:CHM_layer} to the high-dimensional center-pivot convolutions, \ie, center-pivot convolutional Hough matching (CP-CHM).
For position-sensitive isotropic kernel $k_{\text{CP-psi}}$, the two $\frac{n}{2}$D kernels become identical, \ie, $k^{\frac{n}{2}\text{D}}_{c} = k^{\frac{n}{2}\text{D}}_{c'}$, since every weight at the same position within the $\frac{n}{2}$D kernels satisfies $\|\mathbf{p} - \mathbf{x}\|_\mathrm{g} = \|\mathbf{p}' - \mathbf{x}'\|_\mathrm{g}$.
Also, it is interesting to note that if the weights in $k_{\text{psi}}$ and $k_{\text{iso}}$ are sparsified using center-pivot neighbors, the two kernels become identical with the same parameter sharing scheme, \ie, $k_{\text{CP-psi}} = k_{\text{CP-iso}}$.
Without any particular parameter sharing strategies, $k_{\text{CP-full}}$ consists of two different $\frac{n}{2}$ kernels similarly to Eqn.~\ref{eqn:CP_CHM}.
By adapting the proposed parameter sharing and weight-sparsification strategy in high-dimensional kernels, the number of parameters in CP-CHM becomes significantly smaller than that of vanilla CHM kernels;
$k_{\text{CP-psi}}^{\text{6D}}$ and $k_{\text{CP-psi}}^{\text{4D}}$ with $H_\mathrm{k} = W_\mathrm{k} = 5$ and $S_\mathrm{k} = 3$ contain {\em only 12 and 6 parameters} respectively.

Figure~\ref{fig:CHM_kernels} visualizes learned 6D CHM kernels $k_{\text{psi}}^{\mathrm{6D}}$ and $k_{\text{CP-psi}}^{\mathrm{6D}}$ of sizes $H_\mathrm{k} = W_\mathrm{k}=5$ and $S_\mathrm{k}=3$ trained in our experiments.
For the ease of visualizing 6D tensor, we decompose it into multiple (four in case of $k_{\text{psi}}^{\mathrm{6D}}$) 4D tensors in which each of the map shows parameter values of the kernel with the same offset, where the arrows represent the offset vectors relative to the kernel position $(\mathbf{x}, \mathbf{x}')$, and the circles mean zero offset\footnote{Thanks to the reduced number of parameters in center-pivot CHM, all the offset maps are simply merged as one single map for a compact visualization.}.
Figure~\ref{fig:kernel_vis_demo} describes this visualization method in detail. 
The maps reveal that weights for matches with smaller offsets and closer distance are learned to be higher (darker), which appears to be a reasonable voting strategy.




\begin{figure*}
    \begin{center}
        \includegraphics[width=1.0\linewidth]{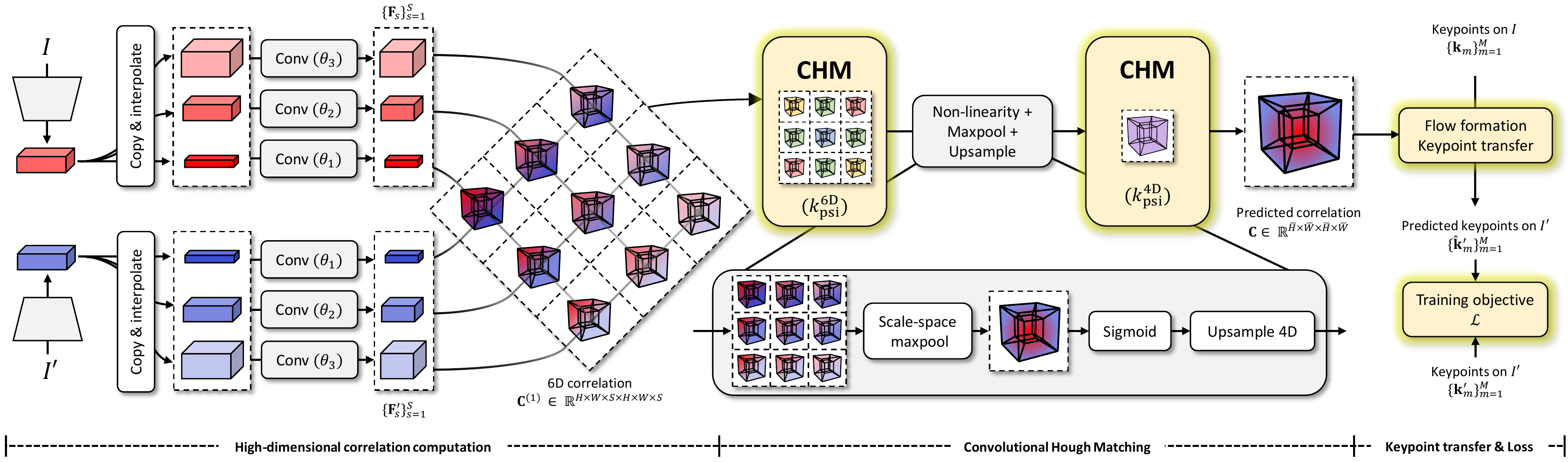}
    \end{center}
    \vspace{-3.0mm} 
      \caption{Overall architecture of the proposed method that performs (learnable) geometric voting in high-dimensional spaces.}
    \vspace{-4.0mm} 
\label{fig:architecture}
\end{figure*}

\section{Convolutional Hough Matching Network}

Based on CHM, we develop a family of image matching models, dubbed {\em Convolutional Hough Matching Networks (CHMNet)}, which consists of three parts: (1) high-dimensional correlation computation, (2) convolutional Hough matching, and (3) flow formation (and keypoint transfer).
Figure~\ref{fig:architecture} illustrates the overall architecture.

\subsection{High-dimensional correlation computation}
Following other recent methods~\cite{huang2019dynamic, li2020correspondence, min2019hyperpixel, min2020dhpf, rocco2018neighbourhood}, we also use as a CNN feature extractor pretrained on ImageNet classification~\cite{deng2009imagenet}.
Given an input image $I$, the feature extractor outputs a feature map in $\mathbb{R}^{C \times H \times W}$.
We construct feature maps of multiple scales $\{{\mathbf{F}}_{s}\}_{s=1}^{S}$ by resizing the output for $S-1$ times by the scaling factor of $\sqrt{2}$, followed by $3 \times 3$ conv layers with parameters $\{\theta_{s}\}_{s=1}^{S}$, reducing channel dimensions of input feature map by $1/\rho$.
The $S$ different conv layers learn to capture effective semantic information of receptive fields with different scales for the subsequent multi-scale (6D) correlation computation.
The same is done for $\{{\mathbf{F}}'_{s}\}_{s=1}^{S}$ given image $I'$.
We set $S=3$, \ie, $\{1/\sqrt{2}, 1, \sqrt{2} \}$, and $\rho=4$ in our experiments. 

Given a set of feature pairs from multiple scales $\{({\mathbf{F}}_{s}, {\mathbf{F}}'_{s})\}_{s=1}^{S}$, we compute all possible 4D correlation tensors placed on the $S \times S$ grid:
\begin{align}
    \mathbf{C}^{(0)}_{mn}(\mathbf{x}_{m}, \mathbf{x}'_{n}) = \text{ReLU}\Bigg(\frac{{\mathbf{F}}_{m}(\mathbf{x}_{m}) \cdot {\mathbf{F}}'_{n}(\mathbf{x}'_{n}) }{\norm{{\mathbf{F}}_{m}(\mathbf{x}_{n})} \norm{{\mathbf{F}}'_{n}(\mathbf{x}'_{n})}} \Bigg), 
\end{align}
where $\mathbf{x}_{m} \in \mathcal{X}_{m}$ and $\mathbf{x}'_{n} \in \mathcal{X}'_{n}$ are spatial positions of feature map at scale $m$ and $n$, respectively, and ReLU clamps negative correlation scores to zero.
To process it in the subsequent 6D CHM layer, we interpolate each 4D correlation $\mathbf{C}^{(0)}_{ij}$ to have the same spatial size to build 6D correlation tensor $\mathbf{C}^{\mathrm{(1)}} \in \mathbb{R}^{H \times W \times S \times H \times W \times S}$ such that ${\mathbf{C}}_{::i::j}^{\mathrm{(1)}} = \zeta_{1}({\mathbf{C}}_{ij}^{\mathrm{(0)}})$
where $\zeta_{1}(\cdot)$ is a function that interpolates input 4D tensor to the size ${H \times W \times H \times W}$.

\subsection{Convolutional Hough Matching}
A CHM layer takes the 6D correlation tensor ${\mathbf{C}}^{\mathrm{(1)}}$ to perform convolutional Hough voting in the space of translation and scaling: 
${\mathbf{C}}^{\mathrm{(2)}} = \text{CHM}({\mathbf{C}}^{\mathrm{(1)}}; k_{\mathrm{psi}}^{\mathrm{6D}})$, where $k_{\mathrm{psi}}^{\mathrm{6D}} \in \mathbb{R}^{H_\mathrm{k} \times W_\mathrm{k} \times S_\mathrm{k} \times H_\mathrm{k} \times W_\mathrm{k} \times S_\mathrm{k}}$ is a 6D position-sensitive isotropic kernel.
In our experiments, we set $H_\mathrm{k} = W_\mathrm{k} = 5$ and $S_\mathrm{k} = 3$ with stride 1 for all dimensions and use zero-padding to the input to retain the same size at the output.
We then perform max-pooling on ${\mathbf{C}}^{\mathrm{(2)}}$ to select the most dominant vote among candidate match scores in the scale space, reducing the tensor dimension to 4D: 
\begin{align}
    {\mathbf{C}}_{ijkl}^{\mathrm{(3)}} = \max_{m,n}{\mathbf{C}}_{ijmkln}^{\mathrm{(2)}}.    
\end{align}
Following recent methods~\cite{min2019hyperpixel, min2020dhpf, jeon2018parn, cho2021semantic, lee2019sfnet}, we employ feature representations from multiple convolutional layers to achieve fine-grained localization.
Specifically, given a pair of image, the backbone network provides two pairs of feature maps from different intermediate layers, \eg, \texttt{conv4\_23} and \texttt{conv5\_3}, similarly to~\cite{lee2019sfnet}.
The two feature pairs are passed to different 6D CHM layers followed by scale-space maxpooling, thus forming two 4D tensors.
These tensors are then merged via element-wise addition to aggregate information of different visual aspects:
\begin{align}
    {\mathbf{C}}^{\mathrm{(3)}} ={\mathbf{C}}^{\mathrm{(3)}}_{\texttt{conv4\_23}} + {\mathbf{C}}^{\mathrm{(3)}}_{\texttt{conv5\_3}}.
\end{align}
We then proceed another CHM with a 4D kernel $k_{\mathrm{psi}}^{\mathrm{4D}} \in \mathbb{R}^{H_\mathrm{k} \times W_\mathrm{k} \times H_\mathrm{k} \times W_\mathrm{k}}$ for additional refinement in translation space.
\begin{align}
    {\mathbf{C}} = \text{CHM}(\zeta_{2}(\sigma({\mathbf{C}}^{\mathrm{(3)}}));k_{\mathrm{psi}}^{\mathrm{4D}}),    
\end{align}
where $\sigma(\cdot)$ is the sigmoid activation function and $\zeta_{2}(\cdot)$ is the upsampling function that resizes input 4D tensor to the size of $\bar{H} \times \bar{W} \times \bar{H} \times \bar{W}$ for fine-grained localization. 
We set $\bar{H} = 2H$ and $\bar{W} = 2W$ in our experiment.


\subsection{Flow formation \& keypoint transfer}

\smallbreak
\noindent \textbf{Flow formation.}
The output ${\mathbf{C}}$ can easily be transformed into a dense flow field by applying kernel soft-argmax~\cite{lee2019sfnet}. We  first normalize the raw correlation scores with softmax:
\begin{align}
    \hat{\mathbf{C}} = \frac{\exp{(\mathbf{G}^{\mathbf{p}}_{kl} {\mathbf{C}}_{ijkl})}} {\sum_{(k',l') \in \bar{H} \times \bar{W}} \exp{(\mathbf{G}^{\mathbf{p}}_{k'l'} {\mathbf{C}}_{ijk'l'})}}, 
\end{align}
where and $\mathbf{G}^{\mathbf{p}} \in \mathbb{R}^{\bar{H} \times \bar{W}}$ is 2-dimensional Gaussian kernel centered on $\mathbf{p} = \argmax_{k,l}{{\mathbf{C}}_{ijkl}}$.  
Using the estimated probability map $\hat{\mathbf{C}}$, we then transfer all the coordinates on dense regular grid $\mathbf{P} \in \mathbb{R}^{\bar{H} \times \bar{W} \times 2}$ of image $I$ to obtain their corresponding coordinates $\hat{\mathbf{P}}' \in \mathbb{R}^{\bar{H} \times \bar{W} \times 2}$ on image $I'$:
$\hat{\mathbf{P}}'_{ij:} = \sum_{(k,l) \in \bar{H} \times \bar{W}} \hat{\mathbf{C}}_{ijkl} \mathbf{P}_{kl:}$.
We now can construct a dense flow field at sub-pixel level using the set of estimated matches $(\mathbf{P}, \hat{\mathbf{P}}')$.

\smallbreak
\noindent \textbf{Keypoint transfer.}
As in~\cite{lee2019sfnet}, one simplest way of assigning a match $\hat{\mathbf{k}}$ to some keypoint $\mathbf{k}=(x_{k}, y_{k})$ is to pick a single, discrete sample of a transferred coordinate such that $\hat{\mathbf{k}} = \hat{\mathbf{P}}_{y_{k}x_{k}}'$.
However, this may cause mis-localized keypoints as the discrete sampling under sub-pixel level hinders fine-grained localization.
To this end, we define a soft sampler $\mathbf{W}^{(\mathbf{k})} \in \mathbb{R}^{\bar{H} \times \bar{W}}$ for given keypoint $\mathbf{k}=(x_{k},y_{k})$ as follows
\begin{align}
    \mathbf{W}_{ij}^{(\mathbf{k})} = \frac{\max{(0, \tau - \sqrt{(x_{k} - j)^{2} + (y_{k} - i)^{2}})}} {\sum_{i'j'} \max{(0, \tau - \sqrt{(x_{k} - j')^{2} + (y_{k} - i')^{2}})}}, 
\end{align}
such that $\sum_{ij}\mathbf{W}_{ij}^{(\mathbf{k})} = 1$ where $\tau$ is a distance threshold.
We assign a match to the keypoint $\mathbf{k}$ by $\hat{\mathbf{k}} = \sum_{(i,j) \in \bar{H} \times \bar{W}} \hat{\mathbf{P}}_{ij:} \mathbf{W}_{ij}^{(\mathbf{k})}$.
The soft sampler $\mathbf{W}^{(\mathbf{k})}$ effectively samples each transferred keypoint $\hat{\mathbf{P}}_{ij}$ by giving weights inversely proportional to the distance to $\mathbf{k}$.

\subsection{Training objective}
We assume that keypoint match annotations are given for each training image pair, as in~\cite{choy2016universal,han2017scnet,li2020correspondence,min2019hyperpixel,min2020dhpf}; each image pair is annotated with a set of coordinate pairs $\mathcal{M}=\{(\mathbf{k}_m, \mathbf{k}'_m)\}_{m=1}^{M}$, where $M$ is the number of annotations. 
Following the aforementioned keypoint transfer scheme, we obtain a set of predicted and ground-truth keypoint pairs on image $I'$: $\{(\hat{\mathbf{k}}_m', \mathbf{k}_m')\}_{m=1}^{M}$ by assigning a match $\hat{\mathbf{k}}_m'$ to each $\mathbf{k}_m$.
Our objective in training is formulated as $\mathcal{L} = \frac{1}{M}\sum_{m=1}^{M}\|\hat{\mathbf{k}}_{m}' - \mathbf{k}_{m}'\|$, which minimizes the average Euclidean distance between the predicted keypoints and the ground-truth ones.



\begin{table*}
        \centering
        \captionof{table}{\label{tab:main_table}Performance on standard benchmarks in accuracy, FLOPs, per-pair inference time, and memory footprint. Subscripts denote backbone networks. Some results are from ~\cite{jeon2018parn,kim2018recurrent, li2020correspondence,liu2020semantic,min2019hyperpixel,min2020dhpf,cho2021semantic}. Numbers in bold indicate the best performance and underlined ones are the second best. Models with an asterisk ($^{*}$) are retrained using keypoint annotations (strong supervision) from~\cite{li2020correspondence}. The first column shows supervisory signals used in training: image-level labels (I), and keypoint matches (K). The second column indicates the use of multi-level backbone features. FLOPs, per-pair inference time and memory footprints are measured on our machine with an Intel i7-7820X and an NVIDIA Titan-XP. The results in last two sections show our results with CP-CHM (top) and CHM (bottom) layers.
        }
        \scalebox{0.9}{
        \begin{tabular}{ccclccccccccccc}
                \toprule
                \multirow{3}{*}{Sup.} & \multirow{3}{*}{\shortstack{data\\aug.}} & \multirow{3}{*}{\shortstack{uses\\multi\\layer?}} & \multirow{3}{*}{Methods} & \multicolumn{2}{c}{SPair-71k} & \multicolumn{2}{c}{PF-PASCAL} & \multicolumn{3}{c}{PF-WILLOW} & \multirow{3}{*}{\shortstack{uses\\nD conv\\kernel?}} & \multirow{3}{*}{\shortstack{FLOPs\\(G)}} & \multirow{3}{*}{\shortstack{time\\({\em ms})}} & \multirow{3}{*}{\shortstack{memory\\(GB)}} \\
                
                 & & & & \multicolumn{2}{c}{PCK @ $\alpha_{\text{bbox}}$} & \multicolumn{2}{c}{PCK @ $\alpha_{\text{img}}$} & \multicolumn{2}{c}{PCK @ $\alpha_{\text{bbox}}$} & PCK @ $\alpha_{\text{bbox-kp}}$ & &  &  & \\ 
                 
                 & & & & 0.1 (F) & 0.1 (T) & 0.05 & 0.1 & 0.05 & 0.1 & 0.1 & & & & \\
                 
                 \midrule
                 
                 \multirow{4}{*}{\shortstack{I}} & - & - & NC-Net$_\textrm{res101}$~\cite{rocco2018neighbourhood} & 20.1 & 26.4 & {54.3} & 78.9 & - & - & 67.0 & 4D & 44.9 & 222 & \underline{1.2} \\
                 
                 & - & - & DCC-Net$_\textrm{res101}$~\cite{huang2019dynamic}  & - & 26.7 & {55.6} & {82.3} & - & - & 73.8 & 4D & 47.1 & 567 & 2.7 \\
                 
                 & - & \cmark & DHPF$_\textrm{res101}$~\cite{min2020dhpf}  & 27.7 & 28.5 & {56.1} & {82.1} & 50.2 & \textbf{80.2} & 74.1 & \xmark & \textbf{2.0} & {58} & 1.6 \\
                 
                 & - & - & PMD$_\textrm{res101}$~\cite{li2021pmdnet} &  26.5 & - & - & 81.2 & - & - & 74.7 & \xmark & - & - & - \\
                 
                 \midrule
                 
                 \multirow{17}{*}{\shortstack{K}} 
                 & - & - & UCN$_\textrm{res101}$~\cite{choy2016universal}          & - & 17.7 & - & 75.1 & - & - & - & \xmark & - & - & - \\
                 
                 & - & \cmark & HPF$_{\textrm{res101}}$~\cite{min2019hyperpixel}      &  28.2 & - & 60.1 & 84.8 & - & - & 74.4 & \xmark & - & 63 & - \\

                 & - & \cmark & SCOT$_\textrm{res101}$~\cite{liu2020semantic} & 35.6 & - & 63.1 & 85.4 & - & - & \textbf{76.0} & \xmark & {6.2} & 151 & 4.6 \\

                 & - & - & SCNet$_{\textrm{res101}}$~\cite{han2017scnet}           & - & - & 36.2 & 72.2 & 38.6 & 70.4 & - & \xmark & - & $>$1000 & - \\
                 
                 & - & \cmark & DHPF$_{\textrm{res101}}$~\cite{min2020dhpf}           & {37.3} & 27.4  & {75.7} & {90.7} & 49.5 & 77.6 & 71.0 & \xmark & \textbf{2.0} & {58} & 1.6 \\
                 
                 & - & - & NC-Net$^\textrm{\textbf{*}}_\textrm{res101}$~\cite{rocco2018neighbourhood} & - & - & - & 81.9 & - & - & - & 4D & 44.9 & 222 & \underline{1.2} \\
                 
                 & - & - & DCC-Net$^\textrm{\textbf{*}}_\textrm{res101}$~\cite{huang2019dynamic} &  - & - & - & 83.7 & - & -  & - & 4D & 47.1 & 567 & 2.7 \\
                 
                 & - & - & ANC-Net$_\textrm{res101}$~\cite{li2020correspondence} &  - & \underline{28.7} & - & 86.1 & - & - & - & 4D & 44.9 & 216 & \textbf{0.9} \\
                 
                 & - & - & PMD$_\textrm{res101}$~\cite{li2021pmdnet} &  37.4 & - & - & 90.7 & - & - & \underline{75.6} & \xmark & - & - & - \\
                 
                 & - & \cmark & PMNC$_\textrm{res101}$~\cite{lee2021patchmatchnc} &  50.4 & - & \underline{82.4} & 90.6 & - & - & - & 4D & - & 960 & 2.6 \\
                 
                 & \cmark & \cmark & CATs$_\textrm{res101}$~\cite{cho2021semantic}   & 49.9 & - & 75.4 & \underline{92.6} & 50.3 & 79.2 & 69.0 & \xmark & - & \underline{35} & 1.6 \\
                 
                 & - & \cmark & CATs$_\textrm{res101}$~\cite{cho2021semantic}   & 43.5 & - & - & - & - & - & - & \xmark & - & \underline{35} & 1.6 \\
                 
                 \cline{2-15} \\[-1.8ex]
                 
                 & - & - & CHMNet$_\textrm{res101}$ (ours)   & {46.3} & \textbf{30.1} & {80.1} & {91.6} & \underline{52.7} & \underline{79.4} & 69.6 & 6D & 19.6 & 54 & 1.6 \\
                 
                 & \cmark & \cmark & CHMNet$_\textrm{res101}$ (ours)   & \underline{51.1} & 26.8 & 80.8 & \textbf{92.9} & \textbf{53.8} & 79.0 & 69.6 & 6D & 21.2 & 62 & 1.8 \\\cline{2-15} \\[-1.8ex]
                 
                 & - & - & CHMNet$_\textrm{res101}^{\text{CP-CHM}}$ (ours)   & 46.2 & 26.5 & 80.1 & 90.7 & 48.9 & 72.6 & 63.4 & 3D+3D & \underline{3.4} & \textbf{32} & 1.5 \\ \\[-1.8ex]
                 
                 & - & \cmark & CHMNet$_\textrm{res101}^{\text{CP-CHM}}$ (ours)   & {47.0} & 27.6 & 81.2 & 91.5 &  {51.9} & 77.0 & 67.3 & 3D+3D & 7.2 & 41 & 1.8 \\\\[-1.8ex]
                 
                 & \cmark & \cmark & CHMNet$_\textrm{res101}^{\text{CP-CHM}}$ (ours)   & \textbf{51.3} & 26.1 & \textbf{83.1} & \textbf{92.9} & \textbf{53.8} & {79.3} & 69.3 & 3D+3D & 7.2 & {43} & 1.8 \\

                \bottomrule
        \end{tabular}
        }
\end{table*}

\section{Experimental Evaluation}
In this section we evaluate the proposed method, compare it with recent state of the arts, and discuss the results.

\smallbreak
\noindent \textbf{Implementation detail.} 
For the feature extractor network, we employ ResNet-101~\cite{he2016deep} backbones, truncated after \texttt{conv4\_23} and \texttt{conv5\_3} layers respectively, pre-trained on ImageNet~\cite{deng2009imagenet}.
Both input and output channel sizes of all the CHM layers are set to 1.
We set spatial size of the input image to $240 \times 240$, thus having $H = W = 15$ and $\bar{H} = \bar{W} = 30$.
Due to parameter sharing structure of $k_{\mathrm{psi}}^{*}$ and $k_{\mathrm{iso}}^{*}$, magnitudes of the loss gradient with respect to the shared weights are unevenly distributed during training time.
To resolve the numerical instability, the shared weights are normalized before the convolution by dividing by the number of times being shared.
The network is implemented in PyTorch~\cite{pytorch} and optimized using Adam~\cite{kingma2015adam} with a learning rate of 1e-3. We finetune the backbone network by setting its learning rate 100 times smaller than CHM layers, \eg, 1e-5.
During training, we apply the same data augmentations used in the recent method of~\cite{cho2021semantic}.

\smallbreak
\noindent \textbf{Datasets.} 
We evaluate the proposed network on three standard benchmark datasets of semantic correspondence: SPair-71k~\cite{min2019spair}, PF-PASCAL~\cite{ham2018proposal}, and PF-WILLOW~\cite{ham2016proposal}.
SPair-71k~\cite{min2019spair} is a highly challenging, large-scale dataset, which contains 70,958 pairs from 18 categories with large variations in view-point and scale.
PF-PASCAL~\cite{ham2018proposal} and PF-WILLOW~\cite{ham2016proposal} respectively contain 1,351 pairs from 20 categories and 900 pairs from 4 categories with small variations in view-point and scale.
Each pair in the datasets consists of keypoint match annotations for semantic parts.


\smallbreak
\noindent \textbf{Evaluation metric.} 
We adopt the standard evaluation metric, percentage of correct keypoints (PCK), for the evaluation.
Given a set of predicted and ground-truth keypoint pairs $\mathcal{K}=\{(\hat{\mathbf{k}}'_{m}, \ \mathbf{k}'_{m})\}_{m=1}^{M}$, PCK is measured by $\mathrm{PCK}(\mathcal{K}) = \frac{1}{M}\sum_{m=1}^{M} \mathbbm{1} [\|\hat{\mathbf{k}}'_{m} - \mathbf{k}'_{m}\| \leq \alpha_{\tau} \cdot \max{(w_{\tau}, h_{\tau})}]$ where $w_{\tau}$ and $h_{\tau}$ are the width and height of either an entire image or an object bounding box, \eg, $\tau \in \{\text{img}, \text{bbox}\}$, and $\alpha_{\tau}$ is a tolerance factor.

\subsection{Results and analysis}
On the SPair-71k dataset, following~\cite{min2019hyperpixel, min2020dhpf}, we evaluate two versions for each model: a finetuned model (F), which is trained on SPair-71k, and a transferred model (T), which is trained on PF-PASCAL.
On PF-PASCAL and PF-WILLOW, following the common evaluation protocol~\cite{choy2016universal, han2017scnet, huang2019dynamic, kim2018recurrent, li2020correspondence, min2019hyperpixel, min2020dhpf, rocco18weak, rocco2018neighbourhood}, our network is trained on the training split of PF-PASCAL~\cite{ham2018proposal} and evaluated on the test splits of PF-PASCAL and PF-WILLOW where the evaluation results on PF-WILLOW is to verify transferability. 
We use the same training, validation, and test splits of PF-PASCAL used in~\cite{han2017scnet}.
The quantitative results are summarized in Tab.~\ref{tab:main_table};
to ensure fair comparisons, we note different levels of supervision, data augmentation used in~\cite{cho2021semantic}, and the use of multi-layer backbone features for each method in the first three columns.
Due to the absence of bounding box annotations in PF-WILLOW, the evaluation threshold of a bounding box, \eg, $\text{max}(w_{\text{bbox}}, h_{\text{bbox}})$, is computed by utilizing keypoint annotations but we found that the previous methods use two different schemes when computing the threshold: $\tau \in \{\text{bbox-kp}, \text{bbox}\}$\footnote{The former ($\text{bbox-kp}$) uses two keypoint positions to approximate the a bounding box that tightly wraps the target object whereas a bounding box of the latter scheme ($\text{bbox}$) loosely covers the object as it uses only a single keypoint position, which typically yields better PCK results.}.
For fair comparisons, we note the two different evaluation thresholds for PF-WILLOW in Tab.~\ref{tab:main_table}: $\alpha_\text{bbox-kp}$ and $\alpha_\text{bbox}$.

\begin{figure}
    \begin{center}
        \includegraphics[width=1.0\linewidth]{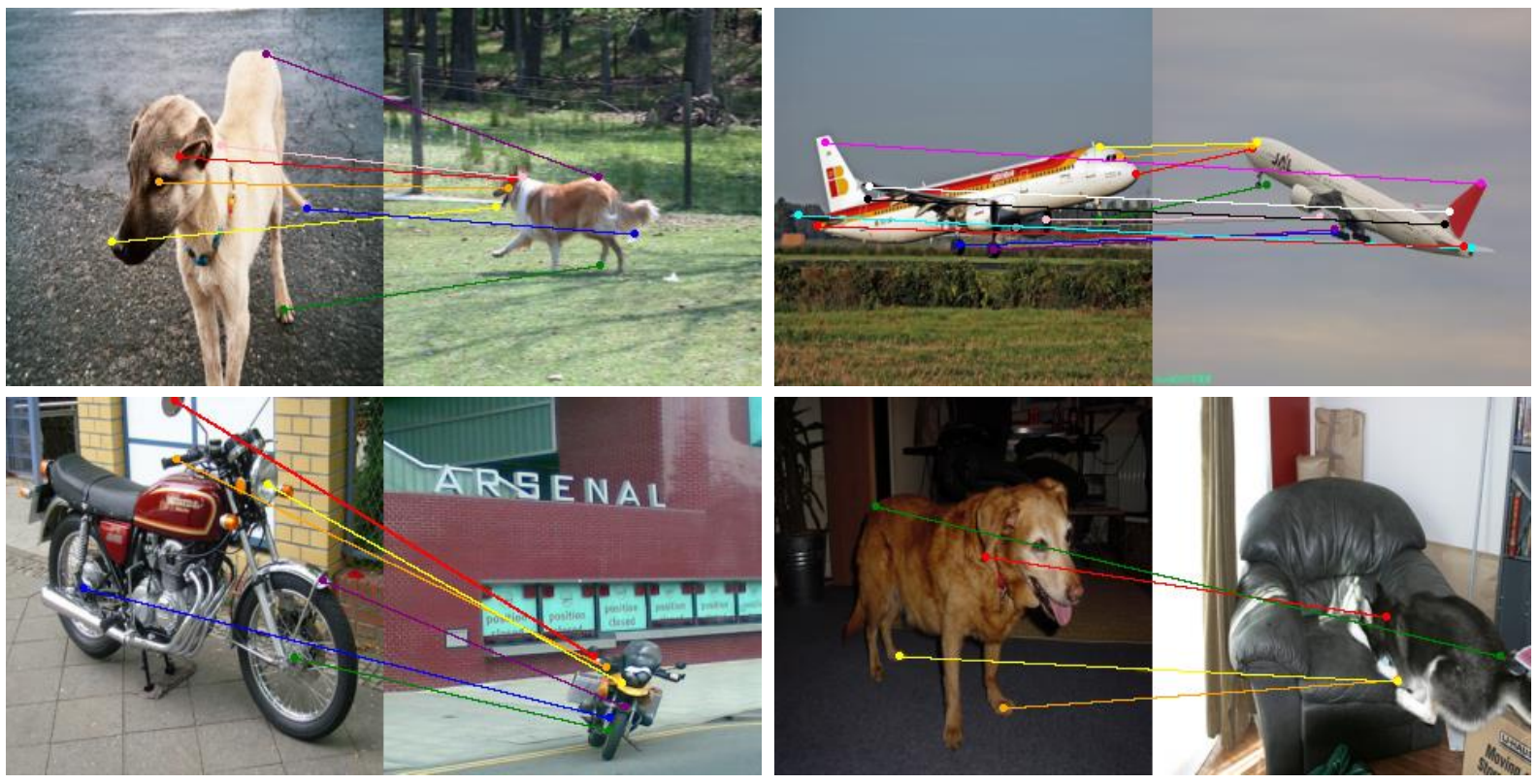}
    \end{center}
    \vspace{-5.0mm} 
      \caption{Qualitative results on SPair-71k dataset. Our model predicts reliable matches under deformations, and large changes in view-point and scale.}
    \vspace{-6.5mm} 
\label{fig:qualitative_spair}
\end{figure}

The proposed model finetuned on SPair-71k (F) clearly surpasses current state of the art, outperforming \cite{lee2021patchmatchnc} by 0.9\%p of PCK ($\alpha_{\mathrm{bbox}}=0.1$) while achieving at most 30 times faster per-pair inference time (960{\em ms} vs. 32-43{\em ms}) and 1.5 times smaller memory footprints (2.6GB vs. 1.5-1.8GB).
On PF-PASCAL, our model achieves 0.7\%p and 0.3\%p improvements with $\alpha_{\mathrm{img}} \in \{0.05, 0.1\}$ over the concurrent work of~\cite{cho2021semantic}.
On PF-WILLOW, we found that strongly-supervised methods generally perform poorly compared to weakly-supervised~\cite{min2020dhpf,li2021pmdnet,huang2019dynamic} and non-finetuning methods~\cite{min2020dhpf,liu2020semantic}.
We conjecture that fitting network parameters using sparsely annotated data, \eg, keypoint matches, is susceptible under domain shifts, harming generalizibility to other datasets~\cite{min2020dhpf,min2019hyperpixel}.
On SPair-71k dataset, using additional backbone features improves our model performance by 0.8\%p of PCK and diverse data augmentation techniques used in~\cite{cho2021semantic} further boosts it by 4.3\%p with only 30 parameters in the matching modules, \eg, center-pivot CHM (CP-CHM) layers\footnote{Respective 6D and 4D CP-CHM layers have 12 and 6 parameters only. Thus the number of parameters in CHM layers amounts to 30 (=12+12+6).}.
Note that our model with CP-CHM layers outperforms the same model with the original CHM layers (bottom section of Tab.~\ref{tab:main_table}) in terms of GFLOPs (3.4 vs. 7.2), time (32{\em ms} vs. 43{\em ms}), memory (1.5GB vs. 1.8GB), the number of parameters in CHM layers (30 vs. 495), and PCK performance (51.3 vs. 51.1 on SPair-71k).
The results clearly support the claim that the neighbors located at relatively insignificant positions in high-dimensional space hinder effective and efficient Hough voting.
Figure~\ref{fig:qualitative_spair} visualizes some example qualitative results on SPair-71k.

\begin{figure*}
    \begin{center}
        \includegraphics[width=1.0\linewidth]{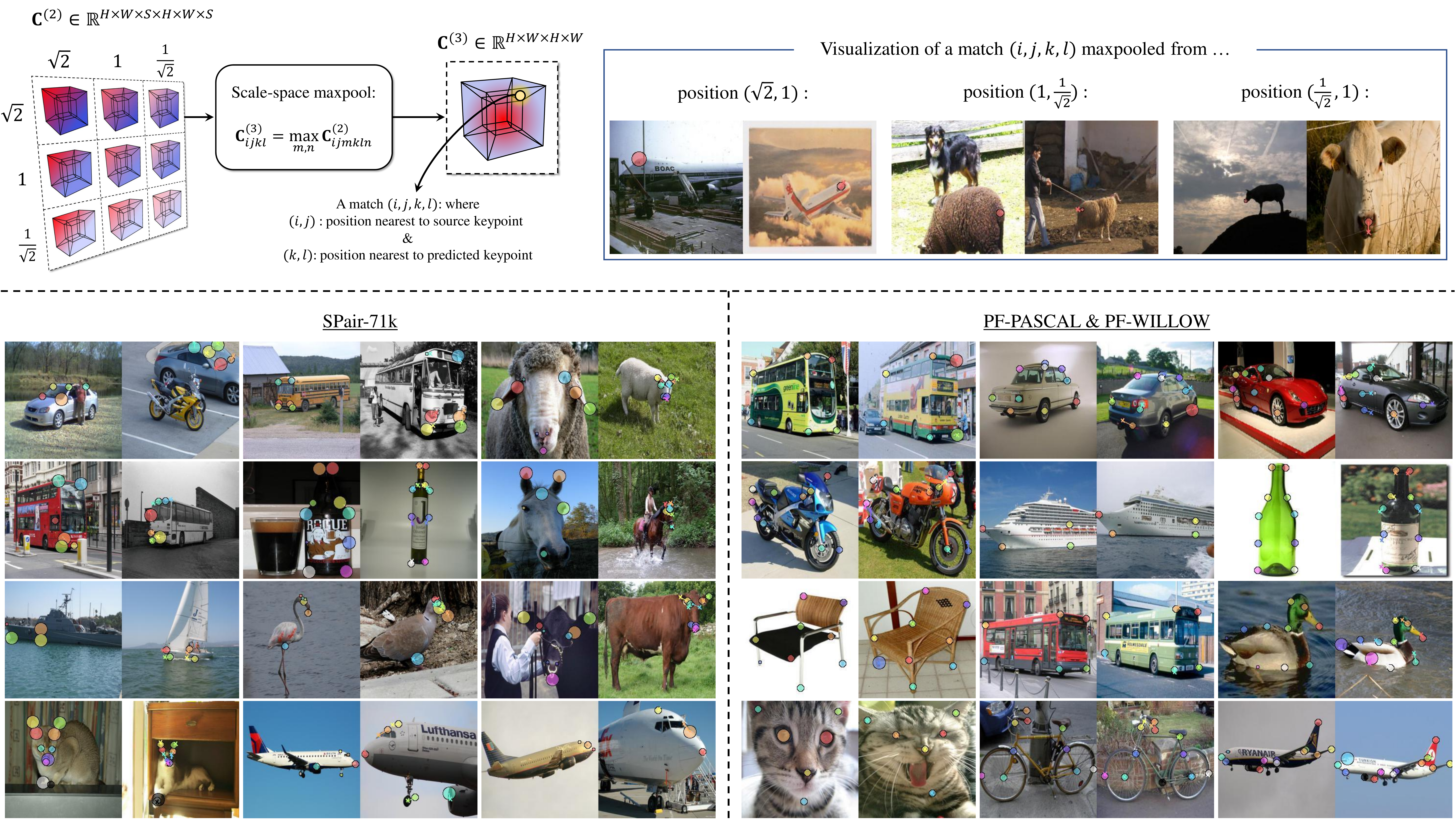}
    \end{center}
    \vspace{-5.0mm} 
       \caption{Visualization of maxpooled position in scale-space. In each image pair, we show source keypoints (given) and their corresponding target keypoints (predicted) in circles in left and right images respectively. The size (large, medium, and small) of each circle indicates maxpooled position in scale-space. If both circles of a match are large, its match score is pooled from position $(\sqrt{2}, \sqrt{2})$ in scale space. If the size of one circle is medium and that of the other is small, its match score is from position $(1, 1/\sqrt{2})$ and so on. We show ground-truth target keypoints in crosses with a line that depicts matching error. Best viewed in electronic form.}
    \vspace{-3.0mm} 
\label{fig:scalemax_qual}
\end{figure*}

\smallbreak
\noindent \textbf{FLOPs, running time, and memory.}
We collect publicly available codes of some recent methods~\cite{huang2019dynamic, li2020correspondence, liu2020semantic, min2020dhpf, rocco2018neighbourhood} to measure their FLOPs, inference time\footnote{Some inference time results are retrieved from~\cite{min2020dhpf}, which is measured on a machine with an Intel i7-7820X and an NVIDIA Titan-XP. For fair comparison, inference time and memory footprint of all the methods are measured on a machine with the same CPU and GPU and includes all the pipelines of a model: from feature extraction to keypoint prediction.}, and memory footprint and compare them with ours in Tab.~\ref{tab:main_table}.
Although the proposed method demands larger memory than some 4D conv based models~\cite{li2020correspondence, rocco2018neighbourhood}, smaller channel sizes of CHM (6D) layers (\{1,1\} vs. \{16,16,1\}) provide noticeable efficiency in terms of GFLOPs (19.6 vs. 44.9).
To achieve faster inference time, we further improve the original implementation of 4D conv~\cite{rocco2018neighbourhood} and develop an efficient nD conv which enables real-time inference (54ms) without increasing FLOPs and memory.
See the supplementary materials for details on our implementation of nD convolution.

\begin{figure}
    \begin{center}
        \includegraphics[width=0.95\linewidth]{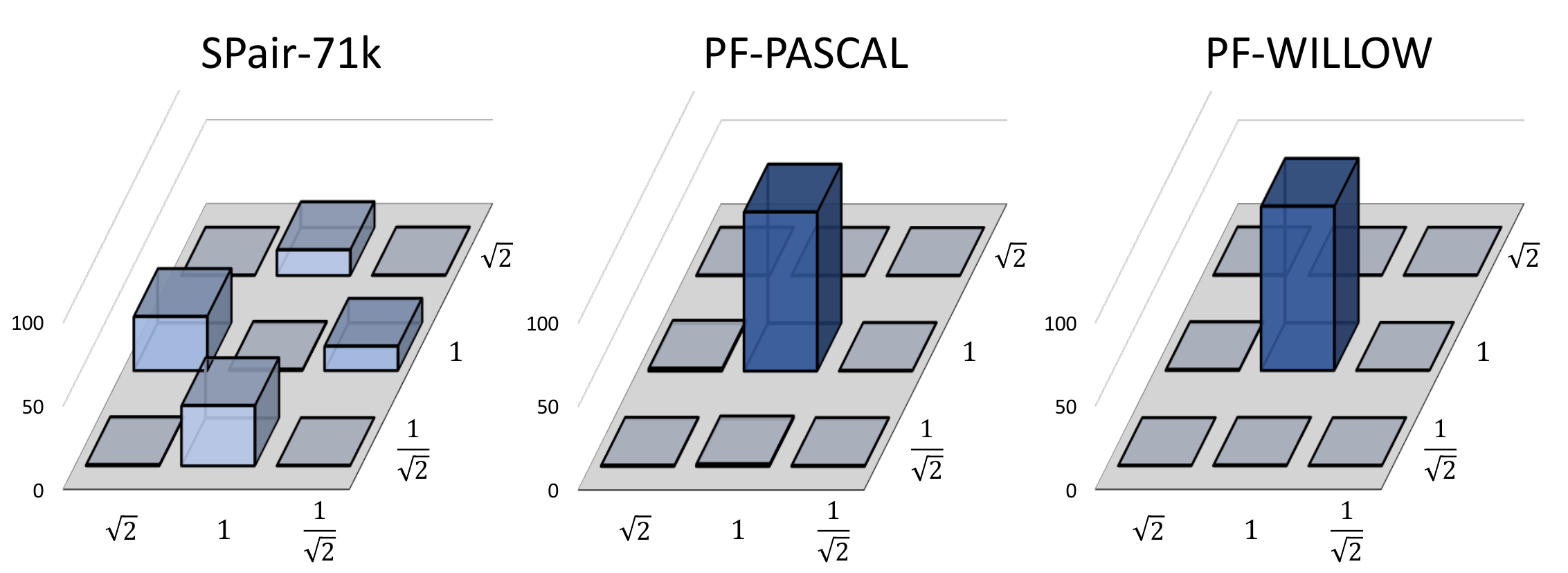}
    \end{center}
    \vspace{-5.0mm}
    \caption{\label{fig:freq_scale} Frequencies over the maxpooled positions in scale-space on SPair-71k, PF-PASCAL, and PF-WILLOW.}
    \vspace{-5.0mm}
\end{figure}

\smallbreak
\noindent\textbf{Analysis on scale-space maxpool.} 
In Figure~\ref{fig:freq_scale}, we also plot frequencies over the maxpooled positions in scale-space after 6D CHM layer ($k_{\mathrm{psi}}^{\text{6D-4D}}$).
The maximum votes on both PF-PASCAL and PF-WILLOW are mostly concentrated on the center scale whereas they are distributed over different scales on SPair-71k;
this is a reasonable voting strategy as objects in PF-PASCAL and PF-WILLOW hardly vary in scale while those in SPair-71k show large variations in both scale and view-point.

To further analyze the results in Fig.~\ref{fig:freq_scale}, we visualize maxpooled positions of predicted matches on sample pairs of SPair-71k~\cite{min2019spair}, PF-PASCAL~\cite{ham2018proposal}, and PF-WILLOW~\cite{ham2016proposal}. Figure~\ref{fig:scalemax_qual} shows the visualization results and describes how we visualize them.
Due to large scale-variations in pairs of SPair-71k, our model collects winners of scale-space vote, \ie, $\text{CHM}(\cdot; k_{\mathrm{psi}}^{\text{6D}})$, from diverse positions in scale-space.
In contrary, objects in PF-PASCAL and PF-WILLOW exhibit relatively small scale-variations, thus encouraging our model to collect winners of the vote mostly from the original scales.
We observe that the maxpooled positions typically depend on scales of object's parts as seen in Fig.~\ref{fig:scalemax_qual}.

\smallbreak
\noindent \textbf{Robustness to background clutter.} 
Recent methods for semantic correspondence~\cite{ham2016proposal, han2017scnet, huang2019dynamic, kim2018recurrent, kim2017dctm, lee2019sfnet, min2019hyperpixel, min2020dhpf, rocco17geocnn, rocco18weak, rocco2018neighbourhood} predict matching scores for all candidate matches but rarely evaluate their robustness to background clutters. 
Here, we compare some recent methods~\cite{li2020correspondence, min2019hyperpixel, min2020dhpf, rocco2018neighbourhood} and ours in terms of robustness to background clutter based on the predicted matching scores.
Each method, however, exploits its correlation tensor differently from others with its own flow formation (keypoint transfer) scheme.
Therefore, given all possible candidate matches in correlation tensor, simply defining matches with top-$k$ scores as positive matches may yield biased estimates.
To ensure fair comparison, for each model, we define a set of coordinates on a regular grid on the input pair of images and assign their best matches using its own keypoint transfer method, thus providing the same number of (fairly collected) candidate matches to every model that we compare.
For each candidate match, we define its match score as a score nearest to spatial position in the correlation tensor.
Given top-$k$ matches according to their matching scores, we define true positives (TPs) as matches falling inside object segmentation masks (bounding box)\footnote{We use object seg. masks and bounding boxes for SPair-71k and PF-PASCAL respectively due to absence of mask annotation in PF-PASCAL.} and false positives (FPs) as those lying outside object masks (boxes).
Precision and recall are measured by  $\frac{N_{\mathrm{TP}}}{N_{\mathrm{TP}} + N_{\mathrm{FP}}}$ and $\frac{N_{\mathrm{TP}}}{N_{\mathrm{mask}}}$, respectively, where $N_{\mathrm{TP}}$ and $N_{\mathrm{FP}}$ are respectively the number of TPs and FPs while $N_{\mathrm{mask}}$ is the number of all candidate matches that fall in the object segmentation masks.
In defining TPs and FPs, we use masks and boxes only due to the absence of dense flow annotation in SPair-71k and PF-PASCAL, but we find that they are good approximation enough to distinguish inliers from outliers in our experimental setup.

\begin{figure}
        \centering
        \includegraphics[width=1.0\linewidth]{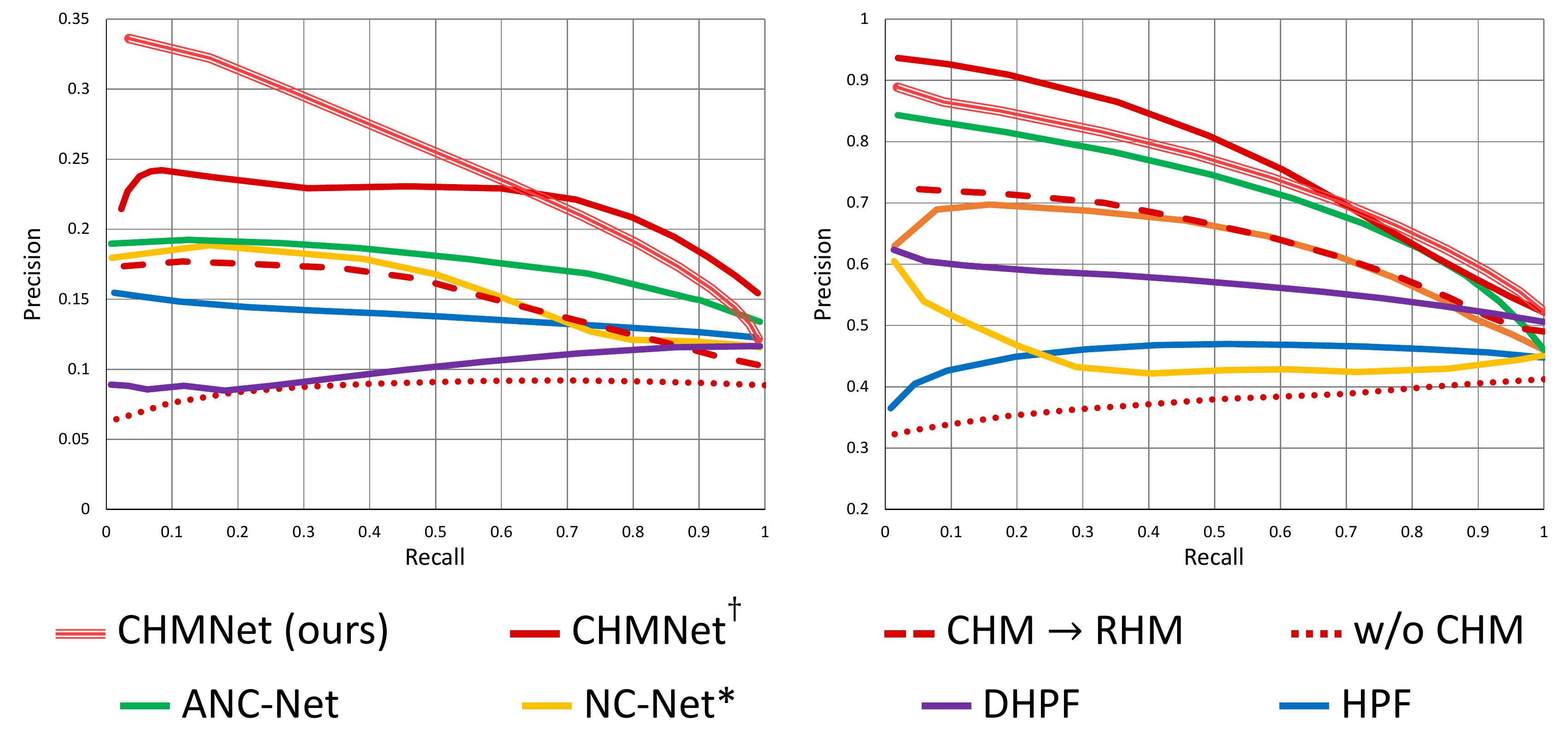}
        \vspace{-6.0mm}
        \caption{\label{fig:pr_curve}PR curves on SPair-71k (top) and PF-PASCAL (bottom). The superscript $\dagger$ denotes our model trained using single-level backbone features.}
    \vspace{-6.0mm}
\end{figure}

Figure~\ref{fig:pr_curve} plots precision-recall curves for the recent methods~\cite{li2020correspondence, min2019hyperpixel, min2020dhpf, rocco2018neighbourhood} and ours. 
The proposed method clearly outperforms other methods, indicating our model effectively discriminates between semantic parts and background clutters as seen in the last row of Fig.~\ref{fig:outlier_qualitative} which visualizes sample pairs with top 300 confident matches.
Given a single-level backbone features as matching primitives (CHMNet$^{\dagger}$), predicted matches become largely unreliable on SPair-71k, assigning low match scores to true matches.
When CHM is either removed (w/o CHM) or replaced with global matching module (CHM $\xrightarrow{}$ RHM), predicted matches become unreliable, being mostly scattered on the background and even hardly regularized.
For our model evaluated on SPair-71k, precision and recall have inverse relationship in most cases.
The superior results of our model on SPair-71k clearly reveal the reliability of our approach under large variations.

\smallbreak
\noindent \textbf{Unbiased evaluation on PF-PASCAL.} 
As discussed in ~\cite{li2020correspondence}, there are overlapping image pairs across the training, validation, and testing splits of the PF-PASCAL dataset.
For unbiased evaluations, we conduct additional experiments on PF-PASCAL after excluding 380 overlapping pairs from the training and validation splits, and summarize the results in Table~\ref{tab:unbiased_pfpascal}.
Our method trained without data augmentations~\cite{cho2021semantic} consistently outperforms other baselines even under the unbiased setting and further improves with additional data augmentations.

\subsection{Ablation study and analysis}

\smallbreak
\noindent \textbf{Analyses on CHM kernel.}
We conduct ablation study on CHM kernel by replacing position-sensitive isotropic kernels with $k_{\mathrm{full}}^{\mathrm{nD}}$\footnote{Note $k_{\mathrm{full}}^{\mathrm{nD}}$ is a n-dimensional kernel without any parameter sharing. The number of parameters in $k_{\mathrm{full}}^{\mathrm{nD}}$ is proportional to $k^{n}$.} and full isotropic ones $k_{\mathrm{iso}}^{\mathrm{nD}}$.
For the ease of notation, we denote by $k_{\mathrm{psi}}^{\text{6D-4D}}$ a model with two CHM layers whose kernels are $k_{\mathrm{psi}}^{\mathrm{6D}}$ and $k_{\mathrm{psi}}^{\mathrm{4D}}$.
In this study, we exclude multi-level backbone features and data augmentations when training the models to focus only on the effect of each CHM kernel.
Table~\ref{tab:ablation_kernel} shows average PCK, its standard deviations, parameter sizes, FLOPs, and average inference time of our model with different kernels over five runs.
Despite a huge difference in the number of parameters (110 vs. 1,250), 
the proposed semi-isotropic kernel $k_{\mathrm{psi}}^{\text{4D-4D}}$ outperforms $k_{\mathrm{full}}^{\text{4D-4D}}$ on Spair-71k (44.5 vs. 43.9) and extending its voting space to 6D, \eg,  $k_{\mathrm{psi}}^{\text{6D-4D}}$, further improves PCK to 46.4 on SPair-71k, which clearly shows efficacy of 6D convolution in scale-space\footnote{To verify the efficacy of the proposed kernel even with sparse match information, we further limit the set of potential matches in $\mathbf{C}^{(0)}$ using $K$ nearest neighbors without using MinkowskiEngine~\cite{choy2019fully} as it does not provide high-dim. kernel customization.
As seen in shaded row in Tab.~\ref{tab:ablation_kernel}, our model with the sparse correlation is comparably effective to $k_{\mathrm{psi}}^{\text{6D-4D}}$, which is consistent to the results of~\cite{rocco2020sparse}. We set $K=10$ in our experiment.}.
The comparable performance of $k_{\mathrm{iso}}^{\text{6D-4D}}$ to $k_{\mathrm{full}}^{\text{6D-4D}}$ reveals that full isotropic parameter sharing can also be a reasonable choice for reducing the large capacity of $k_{\mathrm{full}}^{\text{6D-4D}}$.


\begin{table}[t]
    \begin{center}
    \caption{\label{tab:unbiased_pfpascal} Unbiased evaluation on PF-PASCAL. The results of NC-Net, DCCNet and ANC-Net were taken from ~\cite{li2020correspondence}, where they exclude only 95 overlapping pairs when evaluating. Our method shows better results nonetheless.}
    \vspace{-1.0mm}
    \scalebox{1.0}{
        \begin{tabular}{lcc}
            
                \toprule
                 
                Methods & Original & Unbiased \\
            
                \midrule
                
                HPF$_\textrm{res101}$~\cite{min2019hyperpixel} & 84.8 & 84.2 \\
                SCOT$_\textrm{res101}$~\cite{liu2020semantic} & 85.4 & 86.2 \\
                DHPF$_\textrm{res101}^{\text{weak-sup.}}$~\cite{min2020dhpf} & 82.1 & 79.7 \\
                DHPF$_\textrm{res101}^{\text{strong-sup.}}$~\cite{min2020dhpf} & \underline{90.7} & 84.5 \\
                
                \midrule
                
                NC-Net$_\textrm{res101}$~\cite{rocco2018neighbourhood} & 81.9 & 78.8 \\
                DCCNet$_\textrm{res101}$~\cite{huang2019dynamic} & 83.7 & 78.7 \\
                ANC-Net$_\textrm{res101}$~\cite{li2020correspondence} & 86.1 & 84.2 \\
                
                \midrule
                
                CHMNet$_\textrm{res101}^{\text{CP-CHM}}$ (no aug) & \textbf{92.9} & \underline{87.3} \\\\[-1.8ex]
                CHMNet$_\textrm{res101}^{\text{CP-CHM}}$ (aug) & \textbf{92.9} & \textbf{88.8} \\
                
                \bottomrule
        \end{tabular}}
    \vspace{-0.0mm}
    \end{center}
\end{table}

\begin{table}[t]
    \begin{center}
    \caption{\label{tab:ablation_kernel} Ablation study of CHM kernels over multiple runs.}
    \vspace{-1.0mm}
    \scalebox{0.75}{
        \begin{tabular}{lccccccc}
            
                \toprule
                 
                \multirow{2}{*}{\shortstack{Kernel\\type}} & \multicolumn{2}{c}{SPair-71k PCK ($\alpha_{\text{bbox}}$)} & \multicolumn{2}{c}{PF-PAS. PCK ($\alpha_{\text{img}}$)} & \# params. & FLOPs & time \\
                & 0.05 & 0.1 & 0.05 & 0.1 & in CHM & (G) & ({\em ms}) \\
            
                \midrule
                
                $k_{\mathrm{psi}}^{\text{6D-4D}}$    & \textbf{27.4}$_{\pm 0.16}$ & \textbf{46.4}$_{\pm 0.34}$ & \textbf{80.4}$_{\pm 0.28}$ & \textbf{91.6}$_{\pm 0.23}$ & 275 & \underline{19.6} & 54 \\\\[-1.8ex]
                $k_{\mathrm{full}}^{\text{6D-4D}}$   & 25.9$_{\pm 0.74}$ & 44.8$_{\pm 0.65}$ & \underline{79.8}$_{\pm 0.67}$ & 90.7$_{\pm 0.19}$ & 6,250 & \underline{19.6} & 43 \\\\[-1.8ex]
                $k_{\mathrm{iso}}^{\text{6D-4D}}$    & 24.5$_{\pm 0.28}$ & \underline{44.9}$_{\pm 0.16}$ & 76.5$_{\pm 0.29}$ & 90.2$_{\pm 0.40}$ & \underline{60} & \underline{19.6} & 46 \\
                
                \midrule
                
                $k_{\mathrm{psi}}^{\text{4D-4D}}$    & \underline{26.4}$_{\pm 0.25}$ & 44.5$_{\pm 0.34}$ & 79.3$_{\pm 0.25}$ & \underline{91.1}$_{\pm 0.32}$ & 110 & \textbf{15.9} & 32 \\\\[-1.8ex]
                $k_{\mathrm{full}}^{\text{4D-4D}}$   & 26.1$_{\pm 0.33}$ & 43.9$_{\pm 0.53}$ & 78.4$_{\pm 0.82}$ & 90.3$_{\pm 0.43}$ & 1,250 & \textbf{15.9} & \textbf{26} \\\\[-1.8ex]
                $k_{\mathrm{iso}}^{\text{4D-4D}}$    & 21.0$_{\pm 0.54}$ & 39.7$_{\pm 0.73}$ & 71.8$_{\pm 0.99}$ & 88.0$_{\pm 0.49}$ & \textbf{30} & \textbf{15.9} & \underline{27} \\
                \midrule
                
                \rowcolor{mygray} $k_{\mathrm{psi; sparse}}^{\text{6D-4D}}$    & 26.3$_{\pm 0.18}$ & 45.2$_{\pm 0.41}$ & 80.3$_{\pm 0.86}$ & 91.1$_{\pm 0.05}$ & 275 & - & 55 \\
                \bottomrule
        \end{tabular}}
    \vspace{-4.0mm}
    \end{center}
\end{table}

\begin{figure*}
    \begin{center}
        \includegraphics[width=0.9\linewidth]{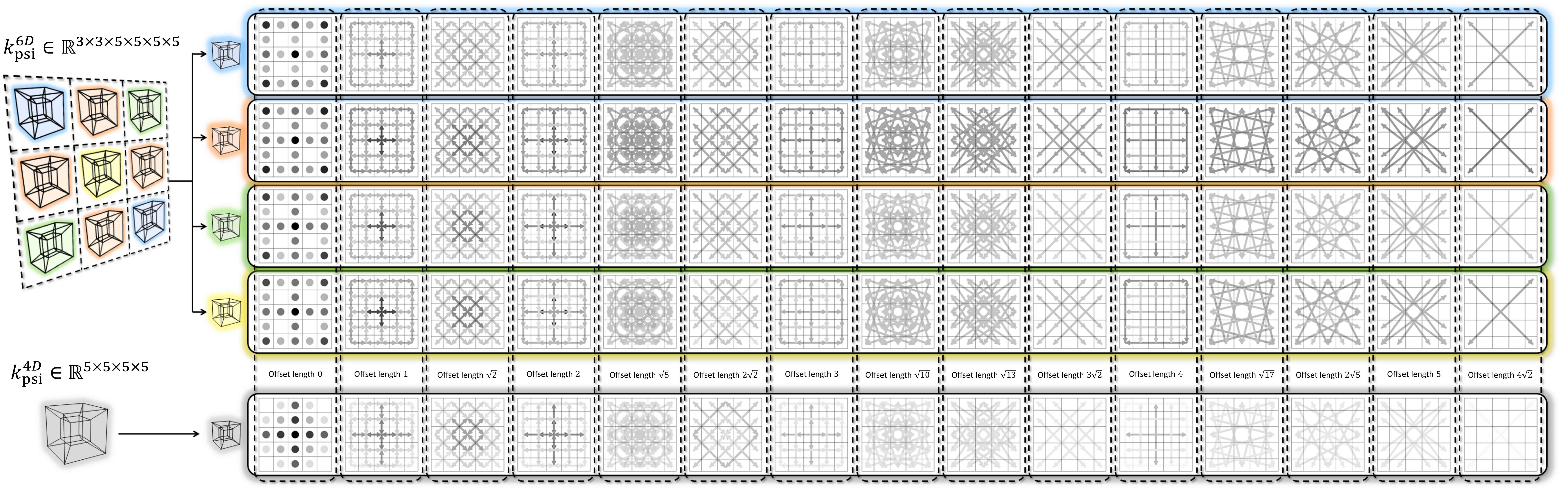}
    \end{center}
    \vspace{-4.0mm} 
      \caption{Learned $k_{\mathrm{psi}}^{\text{6D-4D}}$ used in CHMNet. The 6D kernel ($k_{\mathrm{psi}}^{\text{6D}}$) consists of {\em four} 4D kernels each of which has 55 parameters.}
    \vspace{-1.0mm} 
\label{fig:k_psi}

    \begin{center}
        \includegraphics[width=0.9\linewidth]{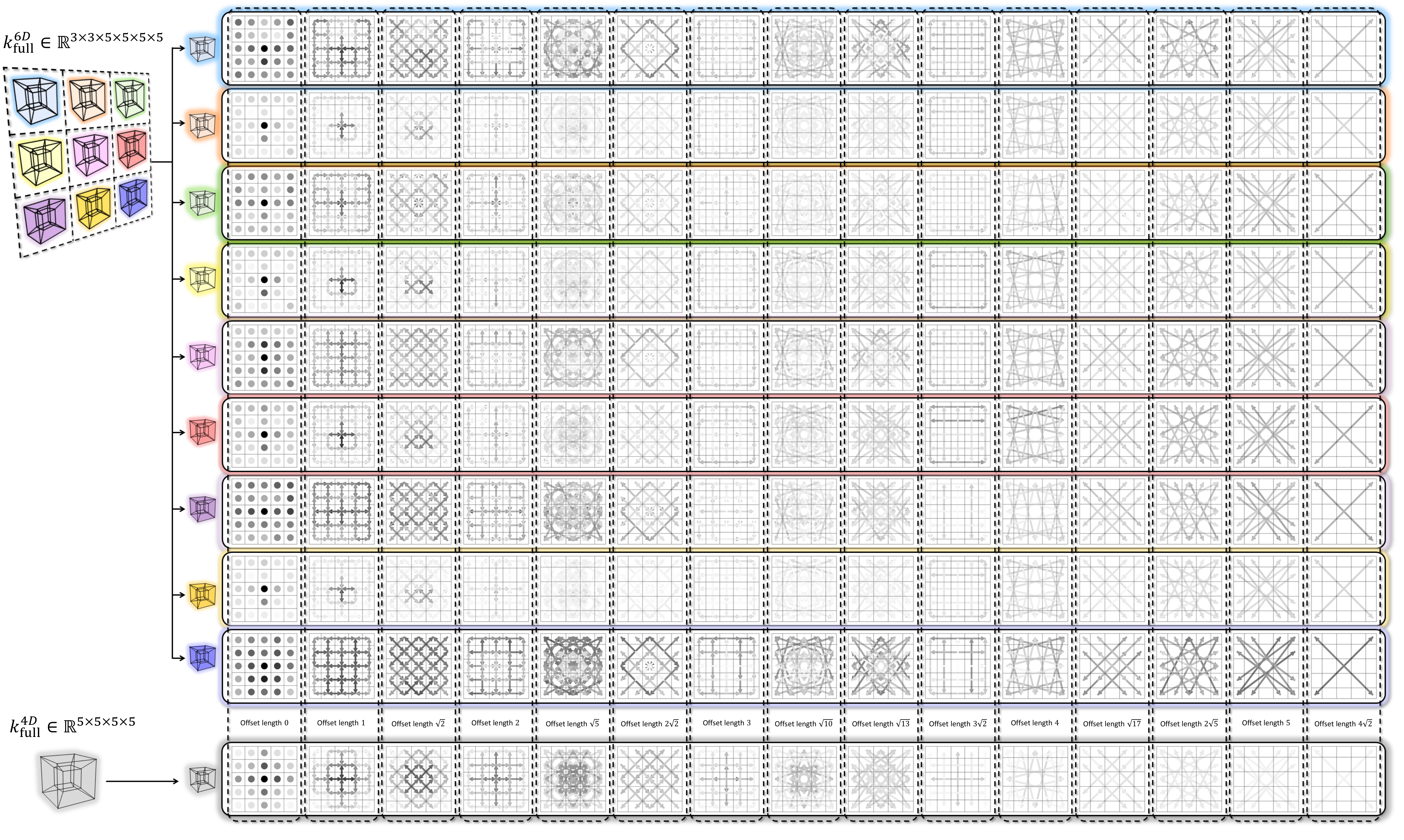}
    \end{center}
    \vspace{-4.0mm} 
      \caption{Learned $k_{\mathrm{full}}^{\text{6D-4D}}$. The 6D kernel ($k_{\mathrm{full}}^{\text{6D}}$) consists of {\em nine} 4D kernels each of which has 625 parameters.}
    \vspace{-1.0mm} 
\label{fig:k_full}

    \begin{center}
        \includegraphics[width=0.9\linewidth]{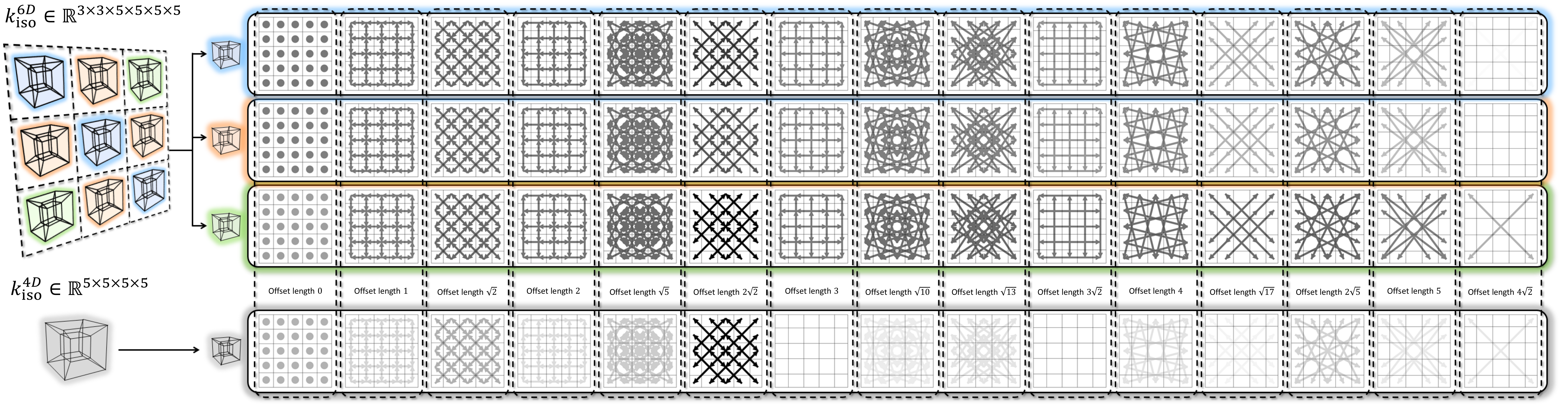}
    \end{center}
    \vspace{-4.0mm} 
      \caption{Learned $k_{\mathrm{iso}}^{\text{6D-4D}}$. The 6D kernel ($k_{\mathrm{iso}}^{\text{6D}}$) consists of {\em three} 4D kernels each of which has 15 parameters.}
    \vspace{-1.0mm} 
\label{fig:k_iso}
\end{figure*}

\begin{figure}[t]
    \begin{center}
        \includegraphics[width=0.95\linewidth]{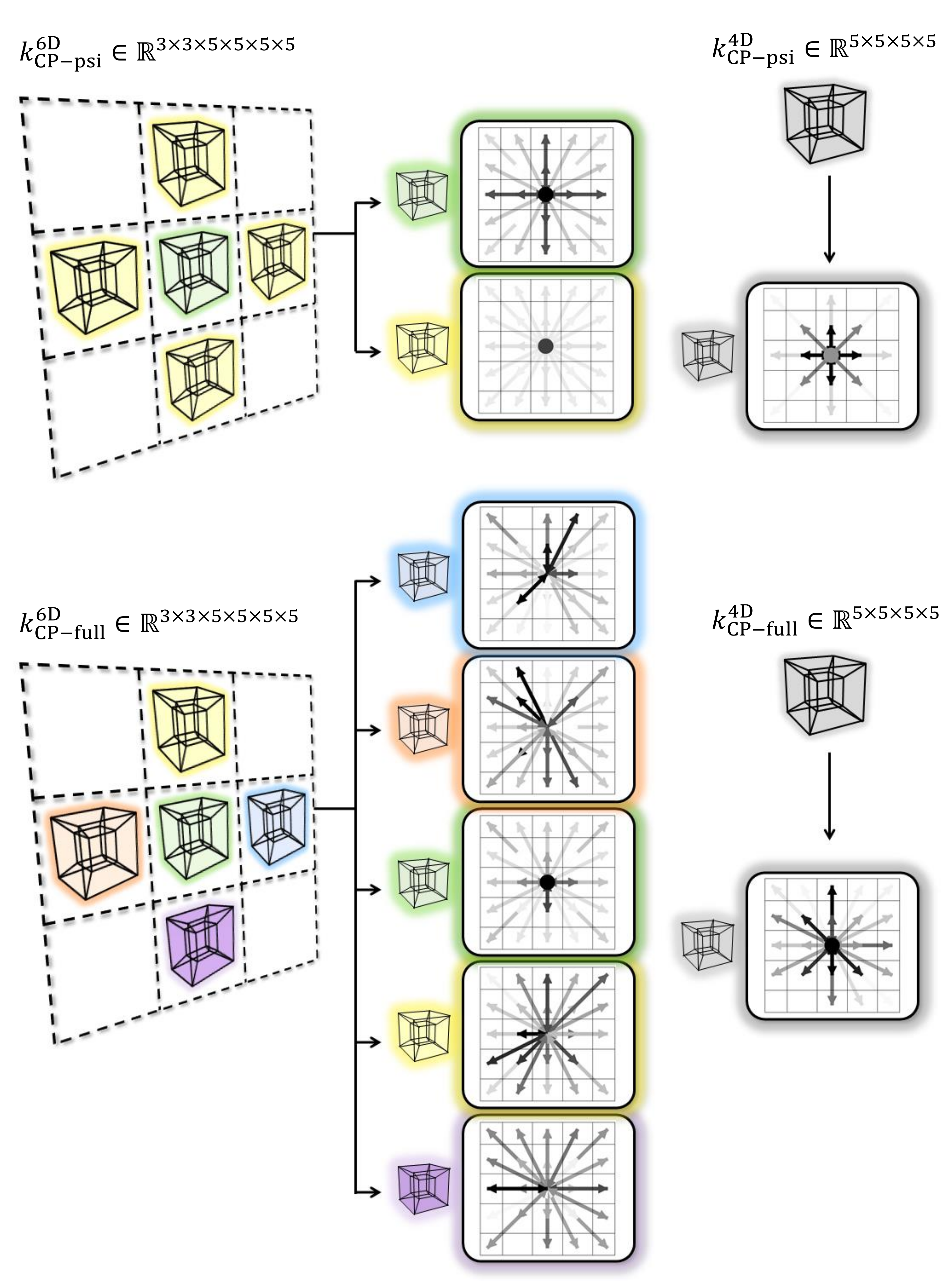}
    \end{center}
    \vspace{-5.0mm} 
      \caption{Visualization of CP-CHM kernels.}
    \vspace{-3.0mm} 
\label{fig:cpchm_visualization}
\end{figure}

\begin{figure}[t]
    \begin{center}
        \includegraphics[width=0.95\linewidth]{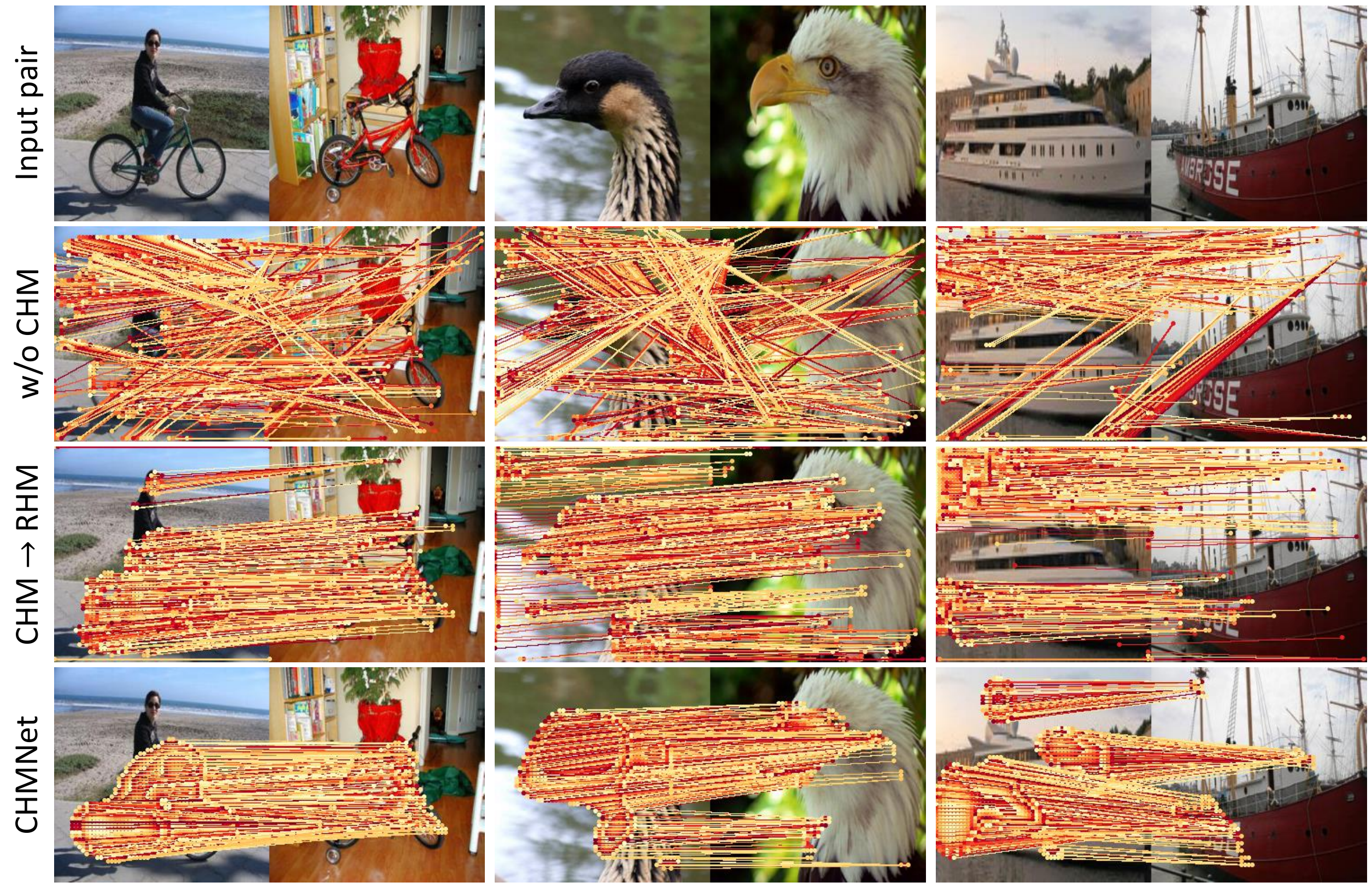}
    \end{center}
    \vspace{-5.0mm}
    \caption{\label{fig:outlier_qualitative}Ablation study on matching modules.}
    \vspace{-0.0mm}

    \begin{center}
        \includegraphics[width=1.0\linewidth]{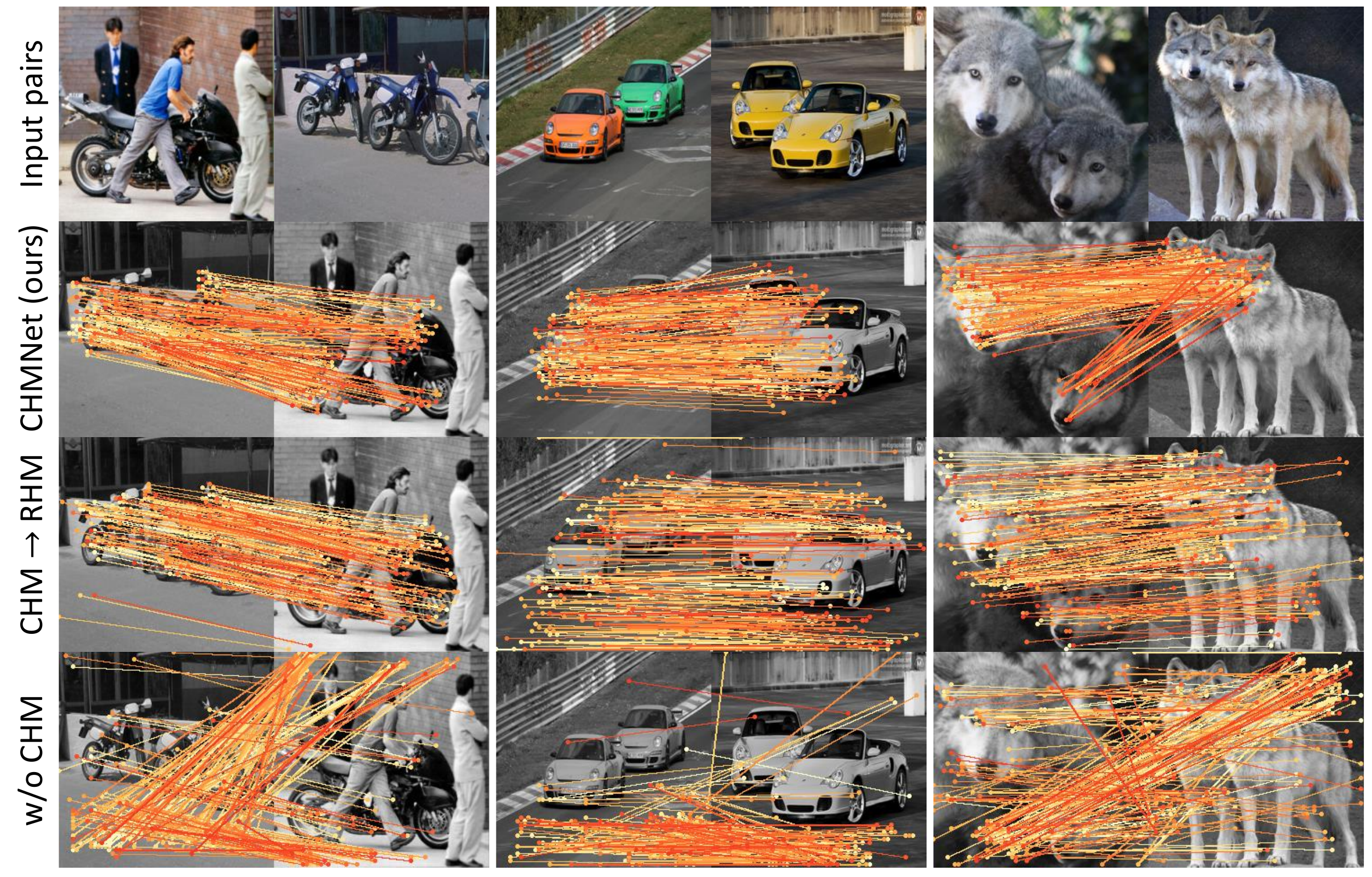}
    \end{center}
    \vspace{-5.0mm} 
       \caption{Multiple instance matching (top 300 confident matches).}
    \vspace{2.0mm} 
\label{fig:multiple}
\end{figure}

\begin{table}[t]
        \centering
        \captionof{table}{\label{tab:scale_space_size}Ablation study of scale-space resolutions.
        }
        \vspace{-2.0mm}
        \scalebox{0.95}{
        \begin{tabular}{cccccc}
                \toprule
                 \multirow{3}{*}{$S$} & \multicolumn{2}{c}{PF-PASCAL} & SPair-71k & \multirow{3}{*}{\shortstack{time\\({\em ms})}} & \multirow{3}{*}{\shortstack{memory\\(GB)}} \\
                
                 & \multicolumn{2}{c}{PCK @ $\alpha_{\text{img}}$} & 
                 PCK @ $\alpha_{\text{bbox}}$ &  &  \\ 
                 
                 & 0.05 & 0.1 & 0.1 (F) && \\
                 
                \midrule
                
                3  & {83.1} & {92.9} & {51.3} & {1.8} & {43} \\
                5 & {82.8} & 92.7 & 51.3 & {2.1} & {64} \\
                7 & 82.4 & {93.0} & {51.6} & 2.2 & 95 \\
                \bottomrule
        \end{tabular}
        \vspace{-15.0mm}
        }
\end{table}

\begin{table}[t]
    \begin{center}
    \caption{\label{tab:ablation_study}Ablation study of core modules in our model.}
    \scalebox{0.9}{
        \begin{tabular}{lcccc}
            
                \toprule
                \multirow{3}{*}{Method} & \multicolumn{2}{c}{SPair-71k} & \multicolumn{2}{c}{PF-PASCAL}  \\
                
                & \multicolumn{2}{c}{PCK ($\alpha_{\text{bbox}}$)} & \multicolumn{2}{c}{PCK ($\alpha_{\text{img}}$)} \\
                 & $0.05$ & $0.1$ & $0.05$ & $0.1$ \\ 
                
                \midrule
                
                CHMNet$_{\mathrm{res101}}$    & 27.2 & 46.3 & 80.1 & 91.6 \\
                
                \midrule
                
                CHM $\xrightarrow{}$ RHM & 21.8 & 38.2 & 77.1 & 89.6 \\
                
                w/o last CHM layer ($k^{\mathrm{4D}}_\mathrm{psi}$) & 24.9 & 43.1 & 79.5 & 89.7  \\
                
                w/o CHM & 10.1 & 21.6 & 61.6 & 78.5  \\
                
                \midrule
                
                w/o kernel $\mathbf{G}$ & 26.6 & 45.5 & 79.5 & 91.3 \\
                
                w/o soft sampler $\mathbf{A^{(\mathbf{k})}}$ & 23.1 & 43.8 & 78.9 & 89.6 \\
                
                \bottomrule
        \end{tabular}}
    \vspace{-4.0mm}
    \end{center}
\end{table}

\begin{figure*}
    \begin{center}
        \includegraphics[width=1.0\linewidth]{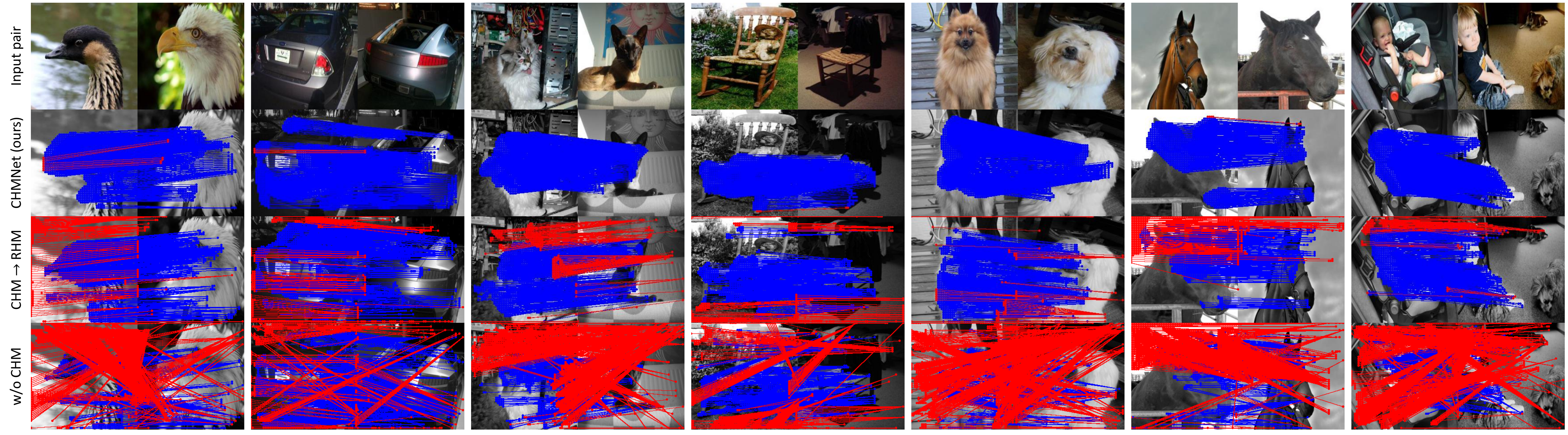}
    \end{center}
    \vspace{-4.0mm} 
       \caption{Sample pairs with top 300 confident matches. TP and FP matches are colored in blue and red respectively.}
    \vspace{-5.0mm} 
\label{fig:pr_qualitative}
\end{figure*}

\begin{figure}[t]
    \begin{center}
        \includegraphics[width=1.0\linewidth]{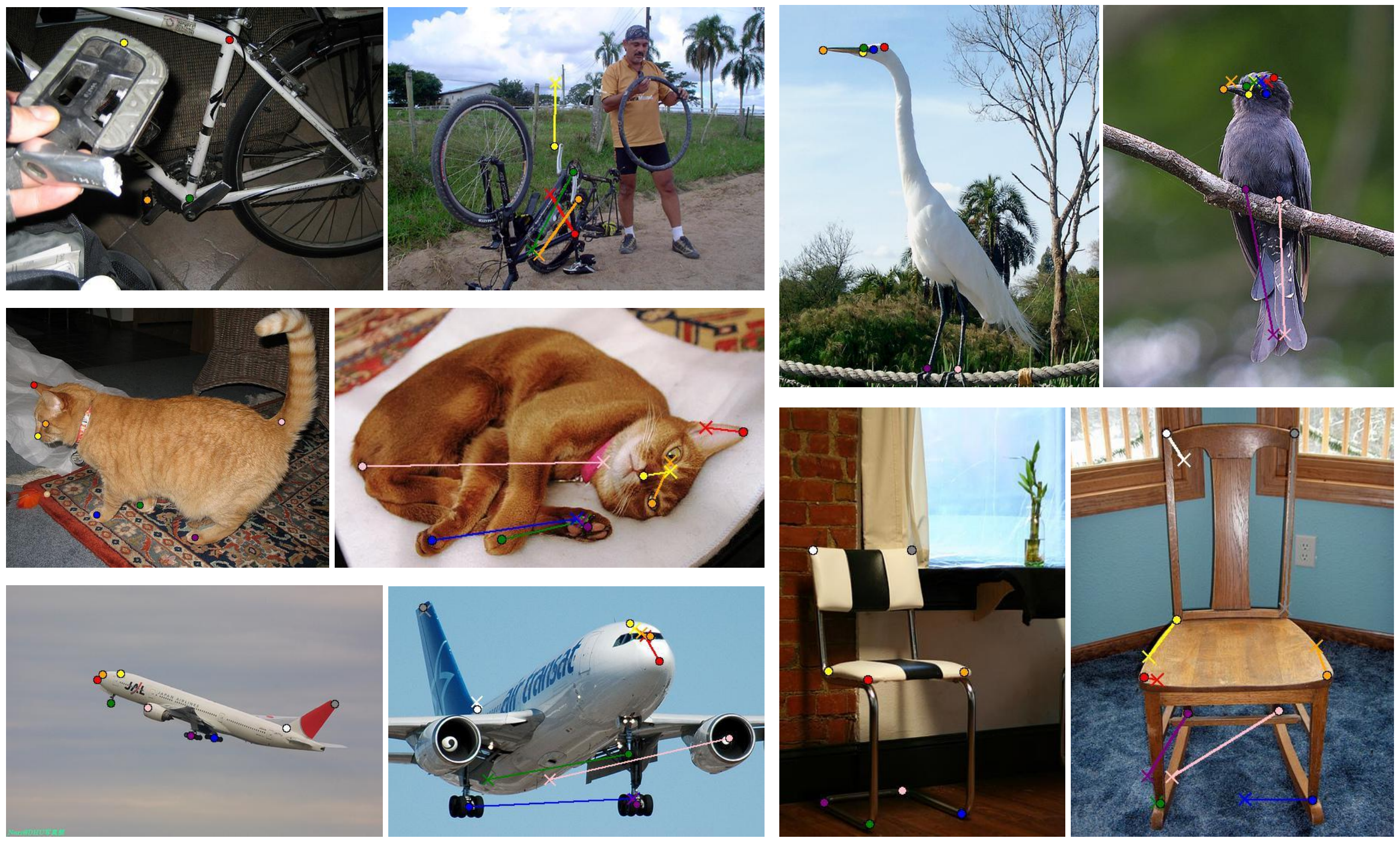}
    \end{center}
    \vspace{-3.0mm} 
       \caption{Failure cases on SPair-71k~\cite{min2019spair} dataset in presence of extreme changes in view-point, large intra-class variation, and deformation. We show the keypoints of ground-truth correspondences in circles and the predicted keypoints in crosses with a line that depicts matching error.}
    \vspace{-5.0mm} 
\label{fig:failure_cases}
\end{figure}

\vspace{3.0mm}
\smallbreak
\noindent \textbf{Visualization of learned CHM kernels.} 
Learned kernels of $k_{\mathrm{psi}}^{\text{6D-4D}}$ (ours), $k_{\mathrm{full}}^{\text{6D-4D}}$, and $k_{\mathrm{iso}}^{\text{6D-4D}}$ are respectively visualized in Figs \ref{fig:k_psi}, \ref{fig:k_full}, and \ref{fig:k_iso}.
Interestingly, the weight patterns of kernels $k_{\mathrm{psi}}^{\text{6D-4D}}$ and $k_{\mathrm{full}}^{\text{6D-4D}}$ are remarkably similar;
the weights for matches with large offsets and closer distance are learned to be higher (darker) while those with small offsets and far distance are learned to be lower (brighter).
Moreover, learned weight patterns of 4D maps in second, fourth, sixth, and eighth rows of $k_{\mathrm{full}}^{\text{6D}}$ in Fig.~\ref{fig:k_full} are noticeably similar to each other.
We also observe that patterns in first and last rows, and patterns in third and seventh rows of $k_{\mathrm{full}}^{\text{6D}}$ are similar to each other as well.
In contrast, $k_{\mathrm{iso}}^{\text{nD}}$ is unable to express diverse weight patterns due to its parameter-sharing constraint that enforces full isotropy.
We also visualize learned center-pivot CHM kernels of $k^{\text{6D-4D}}_{\text{CP-psi}}$ and $k^{\text{6D-4D}}_{\text{CP-full}}$ in Fig.~\ref{fig:cpchm_visualization}.
The learned weights exhibit similar patterns to the those in Figs.~\ref{fig:k_psi}-\ref{fig:k_iso}: the weights with large offsets and near the centers are learned to be higher (darker).
This observation reveals that our kernels $k_{\mathrm{psi}}^{\text{nD}}$ and $k_{\text{CP-psi}}^{\text{nD}}$ in CHMNet clearly benefits from its reasonable parameter-sharing and parameter-sparsification strategies, in terms of both efficiency and accuracy as demonstrated in Tab.~\ref{tab:ablation_kernel}.

\noindent \textbf{Ablation study on size of scale space.}
To study the effect of scale-space resolution of 6D correlation tensor $\mathbf{C}^{\text{(2)}}$, we experiment with three different resolutions: $S \in \{3, 5, 7\}$ and summarize the results in Table~\ref{tab:scale_space_size}. 
From the experiments, we found that increasing the scale-space resolution $>3$ hardly brings noticeable PCK gains while requiring more memory and time during inference. 
We conjecture that such minimal improvement is dataset-related; the image pairs in PF-PASCAL and SPair-71k do not so differ in size that $S=3$ sufficiently covers scale variations in both datasets.

\noindent \textbf{Ablation study on matching modules.}
We analyze the effect of CHM, by either removing or replacing them with the matching module of~\cite{min2019hyperpixel}\footnote{The baseline for this ablation study is our model (CHMNet$_{\text{res101}}$) trained without multi-layer backbone features and data augmentations in order to focus only on the effect of the matching modules.}.
Figure~\ref{fig:outlier_qualitative} and Table~\ref{tab:ablation_study} summarize qualitative and quantitative results, respectively.
The output of global offset voting (CHM $\xrightarrow{}$ RHM) includes many outliers from the background, showing its weakness to the background clutter.
Without the last CHM layer (w/o last CHM), the model fails to effectively refine upsampled correlation scores.
The model prediction is severely damaged without any matching modules (w/o CHM) as seen in second row of Fig.~\ref{fig:outlier_qualitative}.
For keypoint transfer, kernel $\mathbf{G}$ and soft sampler $\mathbf{A}^{(\mathbf{k})}$ help our model find reliable matches by suppressing noisy match scores in $\mathbf{C}$ and effectively aggregating neighborhood transfers, respectively.

The proposed convolutional Hough matching also allows a flexible non-rigid matching and even multiple matching surfaces or objects.
To demonstrate the ability of the CHM in matching multiple objects, we visualize some qualitative results of our method (CHMNet) on some toy images with multiple instances in Fig.~\ref{fig:multiple}.
Top 300 confident matches predicted by our model (CHMNet) are mostly on {\em common} instances in the input pairs of images.
Replacing convolutional Hough matching (learnable local voting layer) to regularized Hough matching~\cite{cho2015unsupervised,min2019hyperpixel} (non-learnable global voting layer) severely damages the model predictions;
the confident matches become noisy and unreliable, mostly being scattered on background.
Without CHM layers, the model fails to localize common instances in the images.
Figure~\ref{fig:pr_qualitative} also visualizes sample pairs of PF-PASCAL with top 300 confident matches predicted by each model.
Our model effectively discriminates between semantic parts and background clutters as seen in the second row of Fig.~\ref{fig:pr_qualitative}.
The absence of CHM layers severely harms the model predictions as seen in the third and last rows of Fig.~\ref{fig:pr_qualitative}.
These results reveal that the proposed CHM layers effectively find reliable matches between common instances across different images while being robust to background clutter even in presence of multiple instances.
Representative failure cases of our model are shown in Fig.~\ref{fig:failure_cases}.

\begin{figure}[t]
    \begin{center}
        \includegraphics[width=0.98\linewidth]{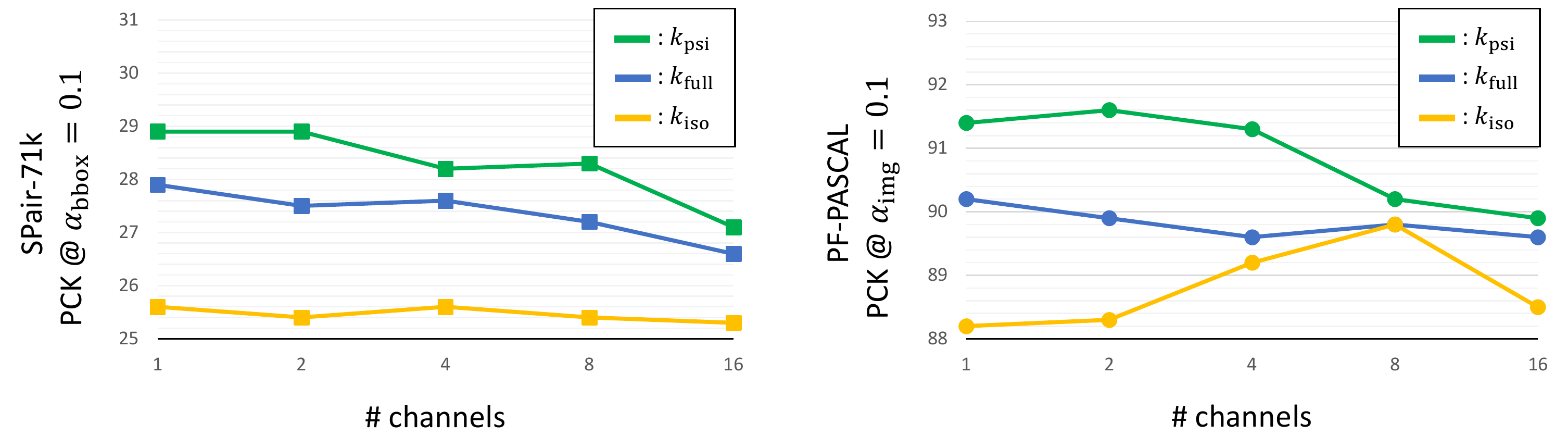}
    \end{center}
    \vspace{-3.0mm}
    \caption{\label{fig:channel_ablation}PCK performance on SPair-71k and PF-PASCAL with different channel sizes of 1, 2, 4, 8, and 16.}
    \vspace{-7.0mm}
\end{figure}

\smallbreak
\noindent \textbf{Effect of channel size.}
To study the effect of channel size, we train our model\footnote{We use the models in the middle section of Tab.~\ref{tab:ablation_kernel}, \eg, $k_{*}^{\text{4D-4D}}$.} using three different kinds of kernels ($k_{\mathrm{psi}}^{\text{4D-4D}}$, $k_{\mathrm{iso}}^{\text{4D-4D}}$, and $k_{\mathrm{full}}^{\text{4D-4D}}$) with different channel sizes, \ie, different number of kernels.
Figure~\ref{fig:channel_ablation} summarizes the results, showing that increasing the channel size rarely brings performance gain and typically harms the quality of prediction for kernels $k_{\mathrm{psi}}^{\text{4D-4D}}$ and $k_{\mathrm{full}}^{\text{4D-4D}}$.
We train the models on the training split of PF-PASCAL and evaluate on test splits of PF-PASCAL and SPair-71k.
For $k_{\mathrm{iso}}^{\text{4D-4D}}$, although increasing channel size improves performance up to certain amount due to its small capacity, it eventually exhibits similar patterns to other kernels after all.

These experiments imply that the high-dimensional convolution on a correlation tensor may play a different role from 2D convolution on an image feature tensor; the role of convolutional matching is to learn a reliable voting strategy rather than to capture diverse patterns in the correlation tensor. This is consistent with the Hough matching perspective, but previous 4D convolution methods~\cite{huang2019dynamic, li2020correspondence, rocco2018neighbourhood, truong2020glunet} with a different perspective commonly use multiple full kernels ($k_{\mathrm{full}}^{\mathrm{4D}}$) for layers.  
To verify our result, we have conducted a similar experiment using the model of~\cite{rocco2018neighbourhood} and obtained the consistent result; the original model, which uses channel sizes of $\{16, 16, 1\}$ for three layers of 4D convolution, achieves 76.2\% PCK on our machine while the model with reduced channels of $\{1, 1, 1\}$ achieves 76.4\% PCK.
Note that in terms of the number of parameters in a layer, our CHM layers ($k^{\text{6D-4D}}_{\mathrm{psi}}$) have $247\sim654$ times smaller number of parameters than the 4D convolution layers used in previous methods~\cite{huang2019dynamic, li2020correspondence, rocco2018neighbourhood, truong2020glunet}. 
This light-weight layer design is particularly important in practice, since the use of multiple channels, \ie kernels, for high-dim convolution quickly increases the cost both in computation and memory.  

For additional results and analyses, we refer the readers to the supplementary.

\vspace{-2.0mm}

\section{Conclusion}
We have introduced the convolutional Hough matching (CHM) and proposed the  powerful matching model, CHMNet, that leverages CHM in a high-dimensional geometric transformation space for establishing reliable visual correspondence.
We also showed that employing center-pivot neighbors to CHM kernels significantly improves model efficiency in terms of both memory and time without harming quality of predictions.
The extensive experiments on several standard benchmarks for semantic visual correspondence demonstrate the benefits of our approach. In particular, our method generalizes existing 4D convolutions and also provides the perspective of Hough transform for geometric matching with interpretable high-dimension kernels. 
We believe further research on this direction can benefit a wide range of other problems related to correspondence. 

\vspace{-3.0mm}

\ifCLASSOPTIONcompsoc
  \section*{Acknowledgments}
\else
  \section*{Acknowledgment}
\fi

This work was supported by Samsung Advanced Institute of Technology (SAIT), the NRF grants (NRF-2017R1E1A1A01077999, NRF-2021R1A2C3012728), and the IITP grant (No.2019-0-01906, AI Graduate School Program - POSTECH) funded by Ministry of Science and ICT, Korea.

\ifCLASSOPTIONcaptionsoff
  \newpage
\fi



\bibliographystyle{IEEEtran}
\bibliography{egbib}

%
\vspace{-40.0mm} 
\begin{IEEEbiography}[{\includegraphics[width=1in,height=1.25in,clip,keepaspectratio]{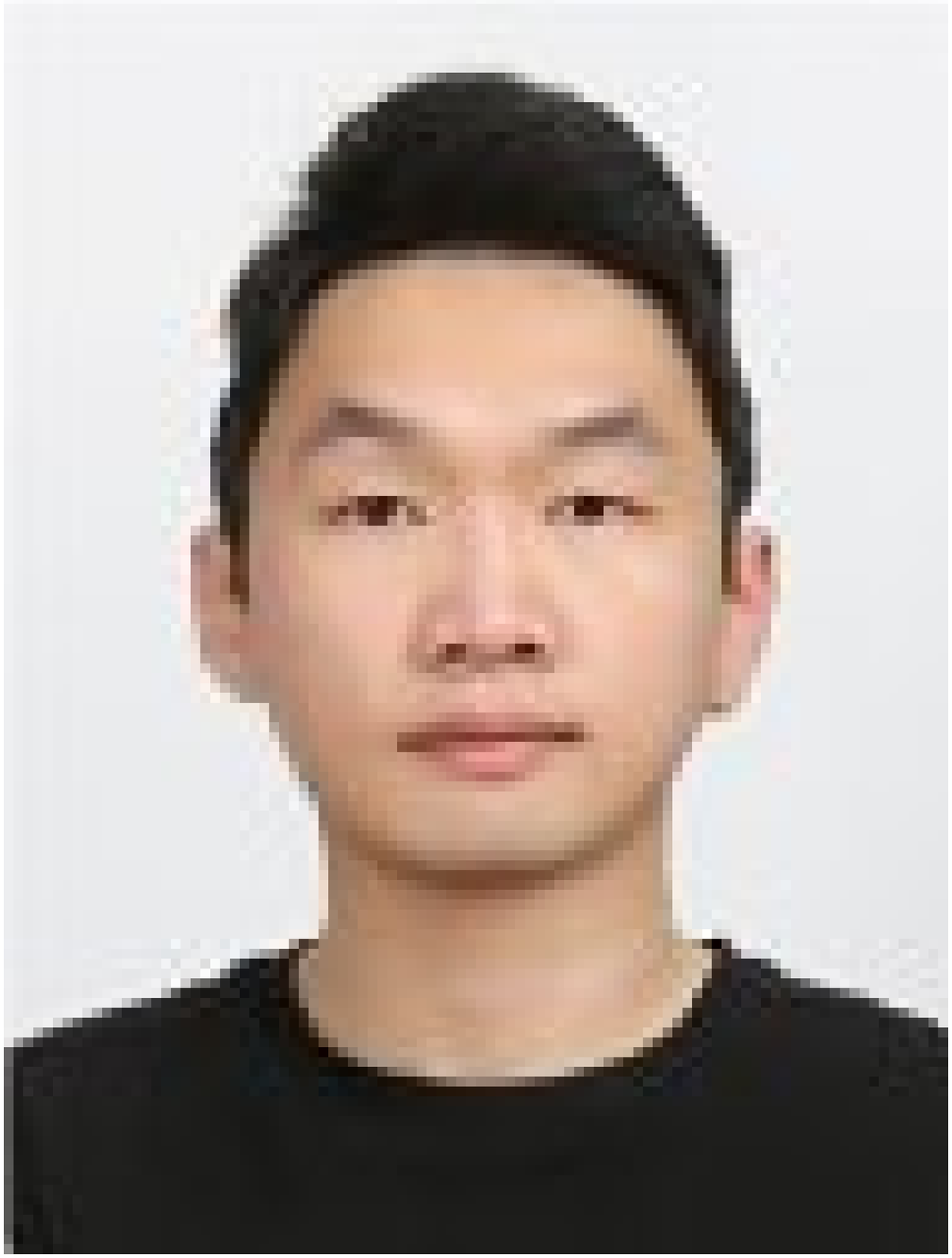}}]{Juhong Min}
received his B.S. degree in Computer Science and Engineering from the Pennsylvania State University in 2014, currently pursuing a Ph.D. degree at POSTECH.
He has received the NAVER PhD Fellowship in 2020 for outstanding research achievements.
His primary research interest lies at learning visual correspondences and its applications such as few-shot learning and visual object tracking.
\end{IEEEbiography}
\vspace{-50.0mm} 
\begin{IEEEbiography}[{\includegraphics[width=1in,height=1.25in,clip,keepaspectratio]{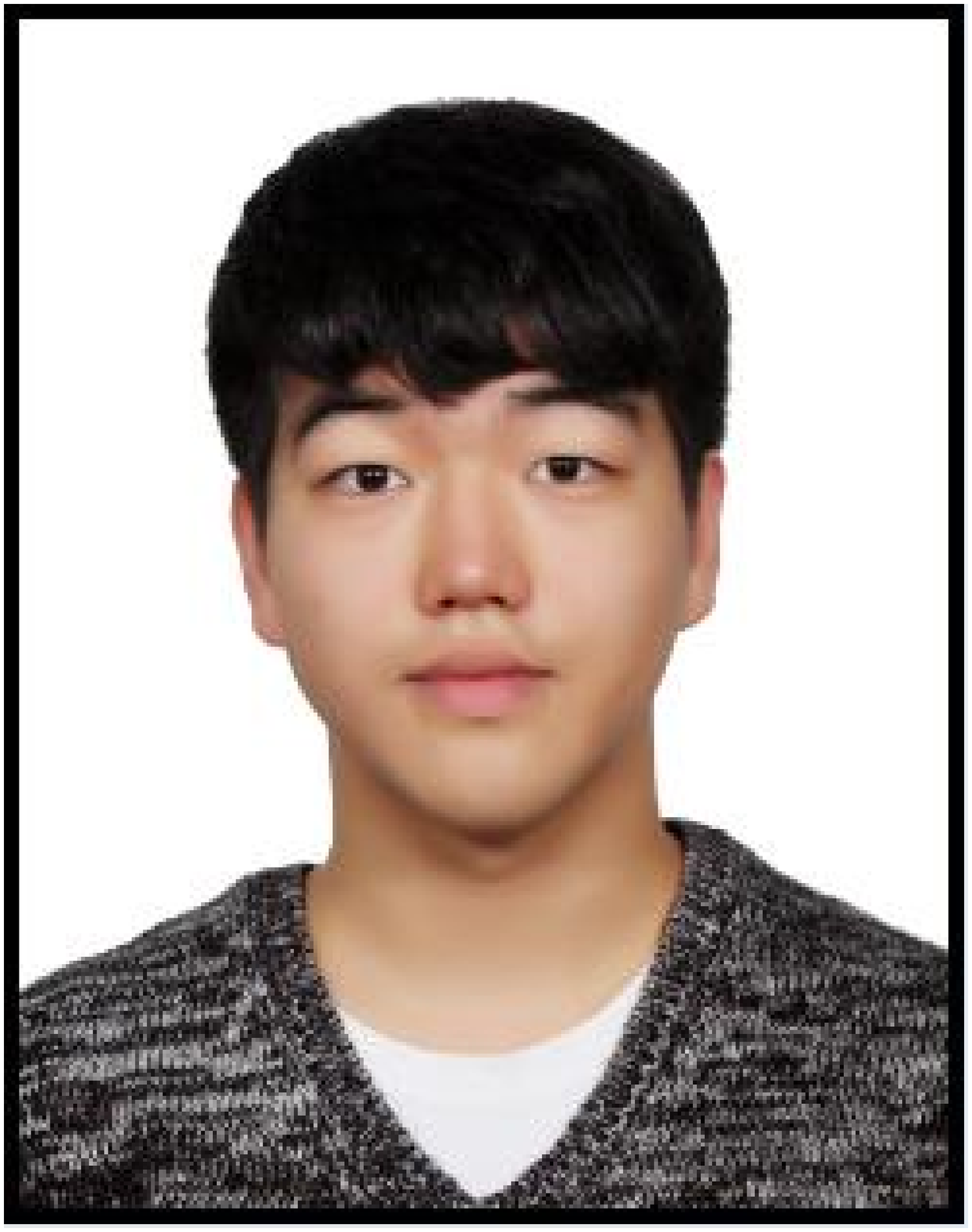}}]{Seungwook Kim}
received his BS degree in Computer Sciences and Engineering from POSTECH in 2020, where he is currently pursuing his PhD degree.
His current research focuses on identifying correspondences between 2D images or 3D point clouds, and applying this idea to potential applications in the computer vision area.
He is also interested in developing a learning-based shape assembly system.
\end{IEEEbiography}
\vspace{-50.0mm}
\begin{IEEEbiography}[{\includegraphics[width=1in,height=1.25in,clip,keepaspectratio]{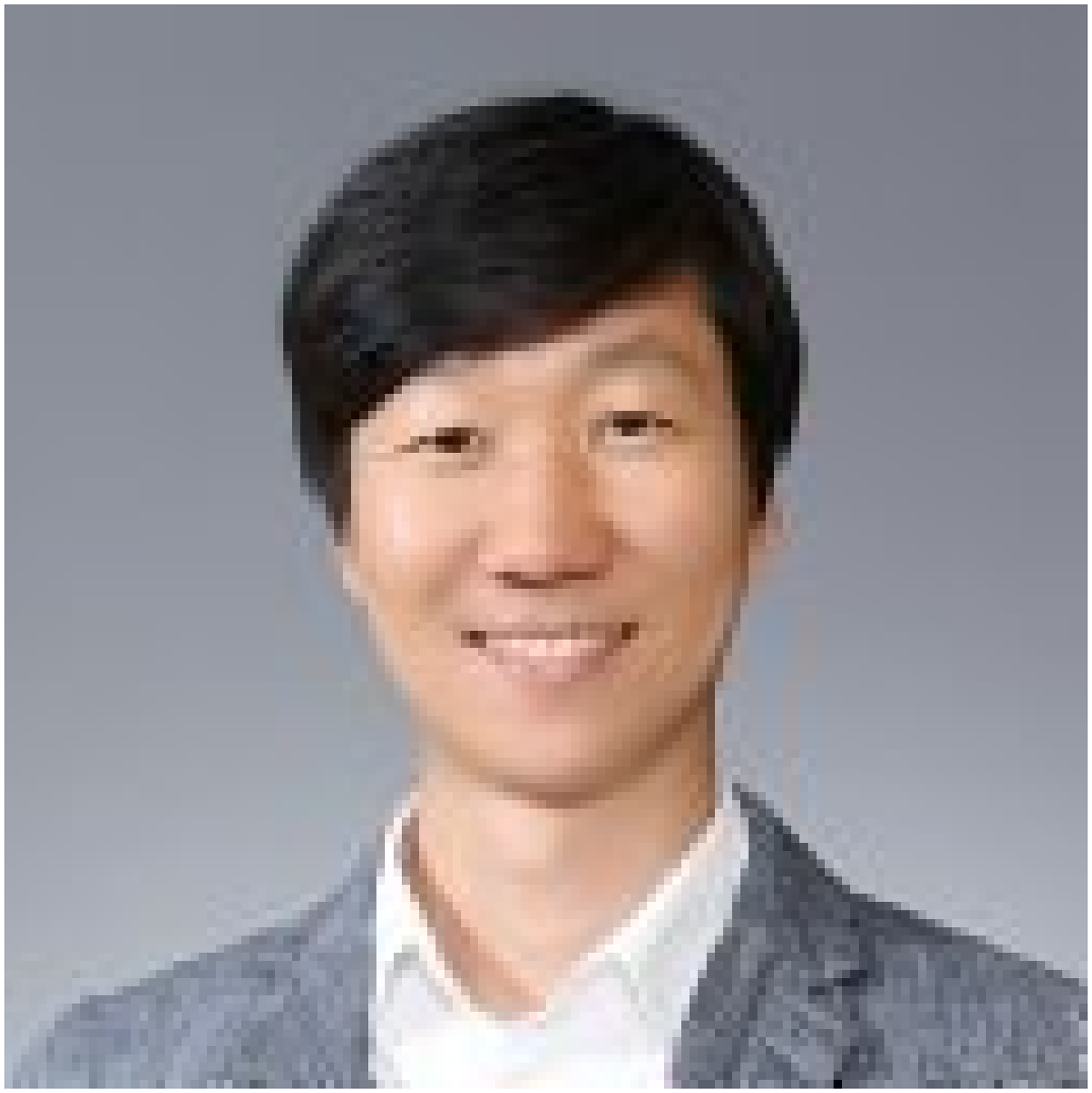}}]{Minsu Cho}
is an Associate Professor of computer science and engineering at POSTECH
in Pohang, South Korea. He obtained his PhD
degree in Electrical Engineering and Computer Science from Seoul National University
in 2012. Before joining POSTECH in 2016, he
worked as an Inria starting researcher in the
ENS/Inria/CNRS Project team WILLOW at cole
Normale Superiure, Paris, France. His research
lies in the areas of computer vision and machine
learning, especially in the problems of object
discovery, weakly-supervised learning, and graph matching
\end{IEEEbiography}




\end{document}